\documentclass[lettersize, journal]{IEEEtran}
\usepackage{cite}
\usepackage{amsmath,amssymb,amsfonts}
\usepackage{algorithmic}
\usepackage{graphicx}
\usepackage{textcomp}
\usepackage[dvipsnames]{xcolor}
\usepackage[font=small,labelfont=bf, center]{caption}
\usepackage[utf8]{inputenc} % allow utf-8 input
\usepackage[T1]{fontenc}    % use 8-bit T1 fonts
\usepackage{hyperref}       % hyperlinks
\usepackage{url}            % simple URL typesetting
\usepackage{booktabs}       % professional-quality tables
\usepackage{nicefrac}       % compact symbols for 1/2, etc.
\usepackage{microtype}      % microtypography
\usepackage{subcaption}
\usepackage{textcomp}
\usepackage{textgreek}
\usepackage{dblfloatfix}
\usepackage{scalerel}
\usepackage{float}
\usepackage{mwe}
\usepackage{verbatim}
\usepackage[numbers,sort&compress]{natbib}
\usepackage{cite}
\usepackage{orcidlink}
\usepackage{ragged2e}

\def\BibTeX{{\rm B\kern-.05em{\sc i\kern-.025em b}\kern-.08em
    T\kern-.1667em\lower.7ex\hbox{E}\kern-.125emX}}

\begin{document}
\title{Domain Adaptation for Underwater Image Enhancement via Content and Style Separation}
\author{Yu-Wei~Chen, and  
Soo-Chang~Pei \orcidlink{0000-0003-2448-4196}, ~\IEEEmembership{Life~Fellow, ~IEEE}
\thanks{Manuscript received Month 00, 2021
This work was supported by the Ministry of Science and Technology, Taiwan,
under Contracts MOST 108-2221-E-002-040-MY3. (Corresponding author:
Soo-Chang Pei.)

    Yu-Wei Chen is with the Graduate Institute of Communication Engineering, National
Taiwan University, Taipei 10617, Taiwan (e-mail: r09942066@ntu.edu.tw).

    Soo-Chang Pei is with the Department of Electrical Engineering, National
Taiwan University, Taipei 10617, Taiwan (e-mail: pei@cc.ee.ntu.edu.tw)}}

\markboth{Journal of \LaTeX\ Class Files,~Vol.~18, No.~9, September~2020}%
{Domain Adaptation for Underwater Image Enhancement via Content and Style Separation}

\maketitle

\begin{abstract}
Underwater image suffer from color cast, low contrast and hazy effect due to light absorption, refraction and scattering, which degraded the high-level application, e.g, object detection and object tracking. Recent learning-based methods demonstrate astonishing performance on underwater image enhancement, however, most of these works use synthetic pair data for supervised learning and ignore the domain gap to real-world data. To solve this problem, we propose a domain adaptation framework for underwater image enhancement via content and style separation, different from prior works of domain adaptation for underwater image enhancement, which target to minimize the latent discrepancy of synthesis and real-world data, we aim to separate encoded feature into content and style latent and distinguish style latent from different domains, i.e. synthesis, real-world underwater and clean domain, and process domain adaptation and image enhancement in latent space. By latent manipulation, our model provide a user interact interface to adjust different enhanced level for continuous change. Experiment on various public real-world underwater benchmarks demonstrate that the proposed framework is capable to perform domain adaptation for underwater image enhancement and outperform various state-of-the-art underwater image enhancement algorithms in quantity and quality. The model and source code will be available at \url{https://github.com/fordevoted/UIESS}
\end{abstract}


\section{Introduction}
\IEEEPARstart{U}{nderwater} image play an important role in marine research, nevertheless, the image taken in underwater are usually affected by color cast, low contrast and hazy effect due to wave-length dependent absorption, light refraction and scattering, which severely degrade the performance of high-level vision algorithms, e.g. object detection and object tracking. To handle this problem, earlier works can divide into two branches, non-physical model based and physical model based. Non-physical model based methods aim to enhance pixel intensity directly to achieve better image quality; Physical model based methods treat the task as an inverse problem, and alleviate the difficulty of solving the ill-posed problem through natural image prior, e.g. Underwater Dark Channel Prior (UDCP) \cite{UDCP}, Generalized Dark Channel Prior (GDCP) \cite{GDCP}, etc.

Recently, deep learning based method demonstrate significant advancement in computer vision and image processing, various works \cite{waterGAN, deepRestoration_SingalLetter2019, underwaterGlobalLocal_Signalprocessing, UWGAN, UGAN, FUnIE-GAN, waternet} are proposed to deal with underwater image enhancement, yet learning based method usually need degraded and high-quality counterpart image pairs for supervised learning, which is infeasible in real world. One 
technical line \cite{waterGAN, FUnIE-GAN, UGAN, NYU_syn} focus on synthesizing realistic underwater image, but the accompanying problem is domain gap between synthetic and real-world images, which most of existing algorithms ignored. Some previous works \cite{UIE-DAL, physical, TUDA} tried to bridge domain gap by domain adaptation, the notion of these works is that minimize discrepancy between encoded latent from synthetic and real-world image domains. In contrast to these works map two domain images into one share latent space, we focus on separate images to domain-variant (style) and domain-invariant (content) latent.   

\begin{figure}[t!]
  \centering
  \begin{subfigure}[b]{0.3\linewidth}
    \centering
    \includegraphics[width=\linewidth]{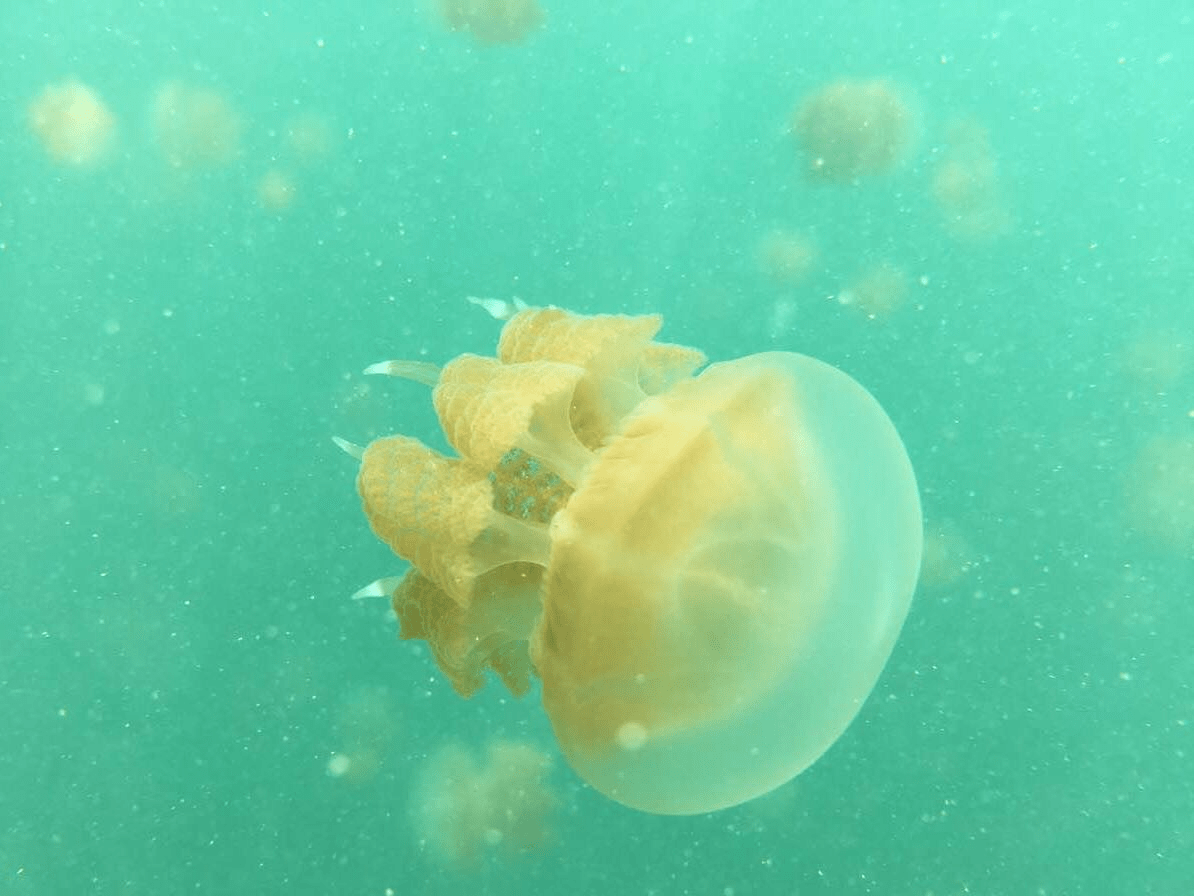}
    \caption{Real underwater image}
    \vspace{-0.32cm}
  \end{subfigure}
  \begin{subfigure}[b]{0.3\linewidth}
    \centering
    \includegraphics[width=\linewidth]{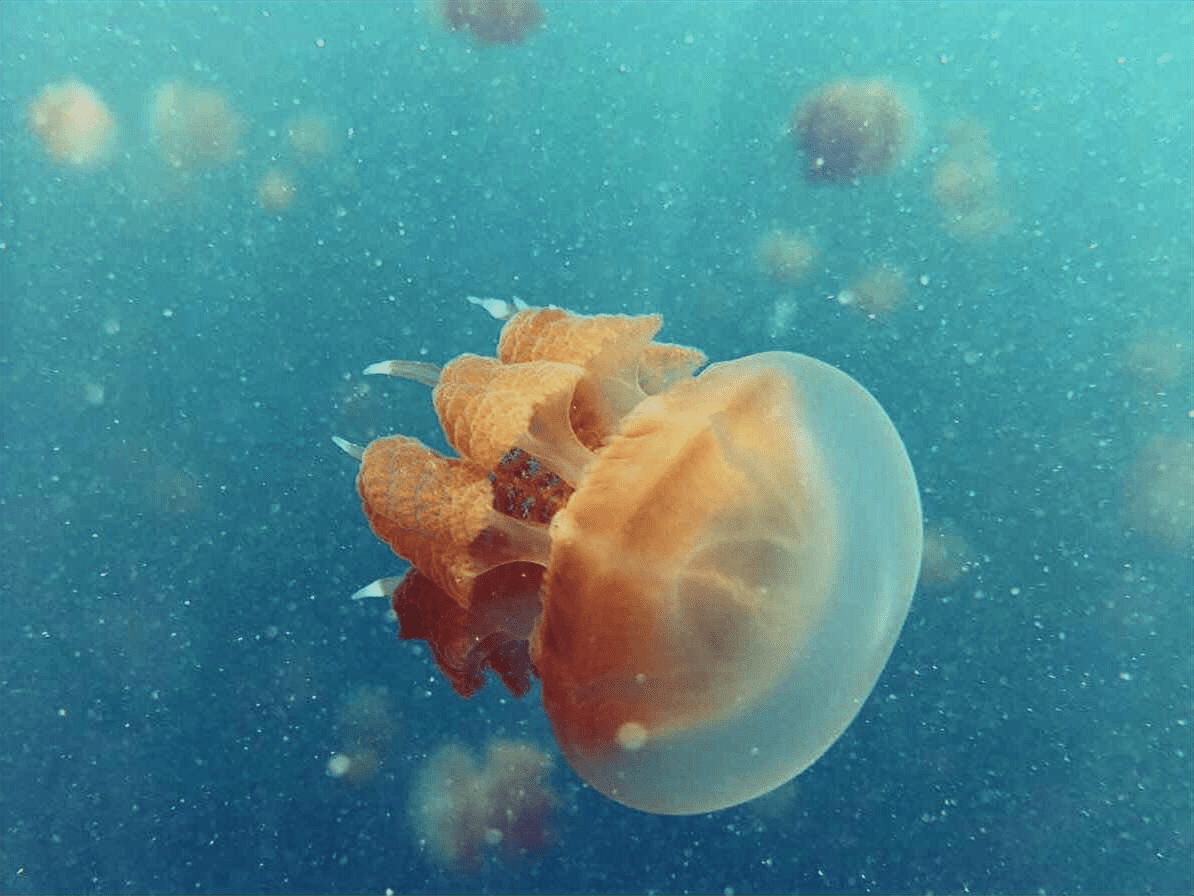}
    \caption{Water-Net \cite{waternet}}
  \end{subfigure}
  \begin{subfigure}[b]{0.3\linewidth}
    \centering
    \includegraphics[width=\linewidth]{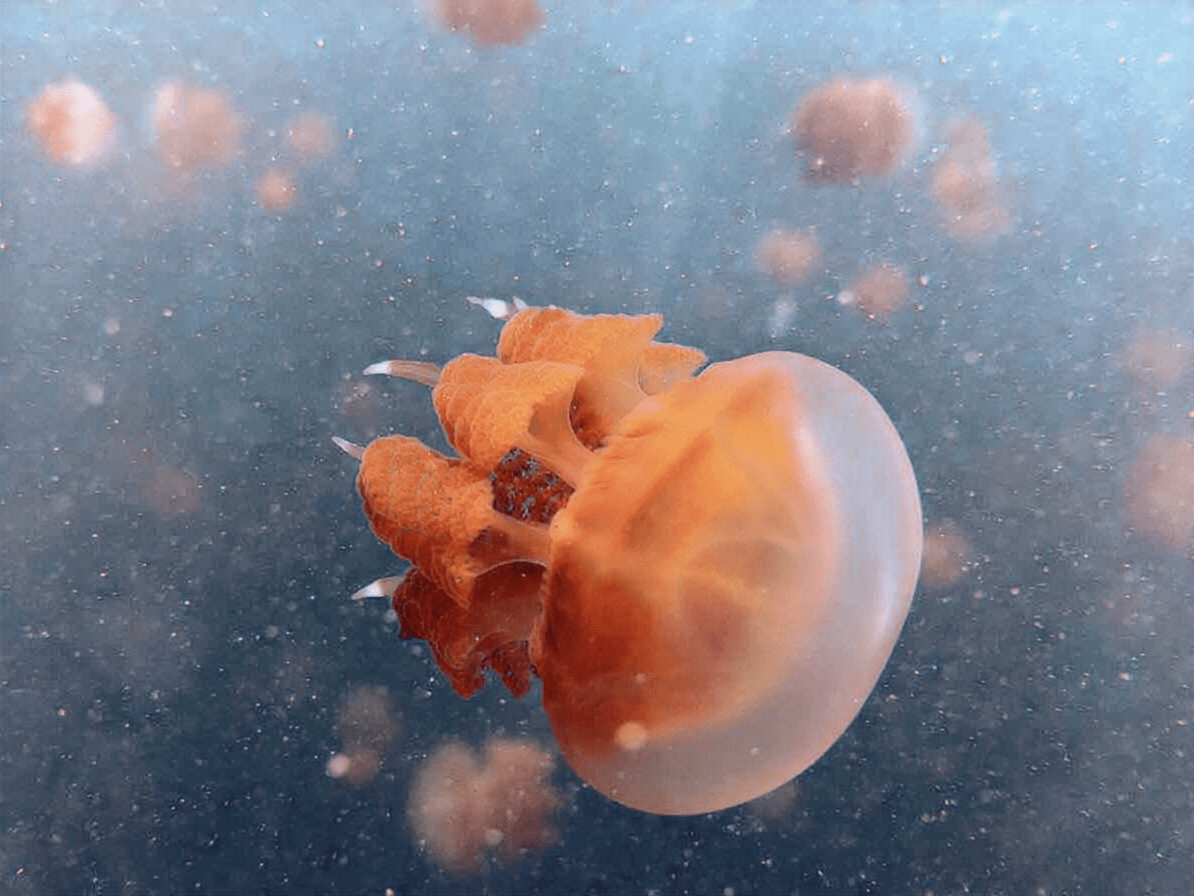}
    \caption{UIESS (Ours)}
  \end{subfigure}
  \vspace{0.4cm}
  \captionsetup{justification=raggedright,singlelinecheck=true}
  \caption{Underwater image enhancement result on real-world image. Proposed methods can generate clearer, higher saturation, and more visual pleasing result.}
\label{fig_demo}
\end{figure}  

In this paper, we proposed a novel domain adaptation framework for underwater image enhancement via content and style separation, the framework contain one share content encoder, two style encoder for synthetic and real-world underwater image, a latent transform unit, a generator and a discriminator. We first separate input underwater image into content and style representation, then based on the assumption of encoded style latent of synthetic, real-world underwater and clean image belongs to three different domains in style latent space, the latent transform unit is responsible to transform degraded style latent into clean one, and the generator generated corresponding image by feeding content and specific style latent. The generator is enforced to distinguish style from different domains by given appropriate constraints. To solve the problem of lacking ground truth of real image, we generate pseudo real image pair, which is composed by syn2real underwater image and high-quality counterpart of synthesis image, to perform domain adaptation. Though the notion of generating pseudo real image pair to perform domain adaptation is common, we argue that disentangled image into content and style is more effective manner than encode domain-agnostic latent from different domain \cite{UIE-DAL, physical, TUDA} or without latent space assumption \cite{DAdehazing, two_step}, we will demonstrate the effectiveness in Section \ref{section_exp}. Our model learn cross domain image-to-image translation and enhancement simultaneously, and can carry out style latent manipulation to obtain the continuous change from input image to enhanced image, the detail would be elaborated in Section \ref{section_method}. As Fig \ref{fig_demo} shown, our model can produce more visual pleasing result compare with Water-Net \cite{waternet}, which is proposed in recent years. Experiment on various public real-world underwater datasets show that our network outperform various state-of-the-art underwater image enhancement algorithms in quantity and quality. 
   
    The summary of our contributions are shown below:
    \begin{itemize}
    \item We propose a novel domain adaptation framework for underwater image enhancement via content and style separation, called UIESS, to bridge the domain gap between synthesis and real-world underwater image. To the best of our knowledge, this is the first attempt to leverage content style disentangling on domain adaptation for underwater image enhancement. Proposed framework can perform cross-domain image-to-image translation and enhancement simultaneously.  
    \item The proposed framework can perform latent manipulation using original style latent and enhanced one to obtain continuous change of different enhancement level, which can act as user interact parameter and manipulate the enhanced result.
    \item Compare with off-the-shelf underwater image enhancement algorithms, experiment on various public underwater real-world and synthesis datasets show that proposed method is superior than current methods on synthesis and real-world datasets.
    
    \end{itemize} 

\begin{figure*}[t!]
  \centering
    \includegraphics[width=0.8\textwidth]{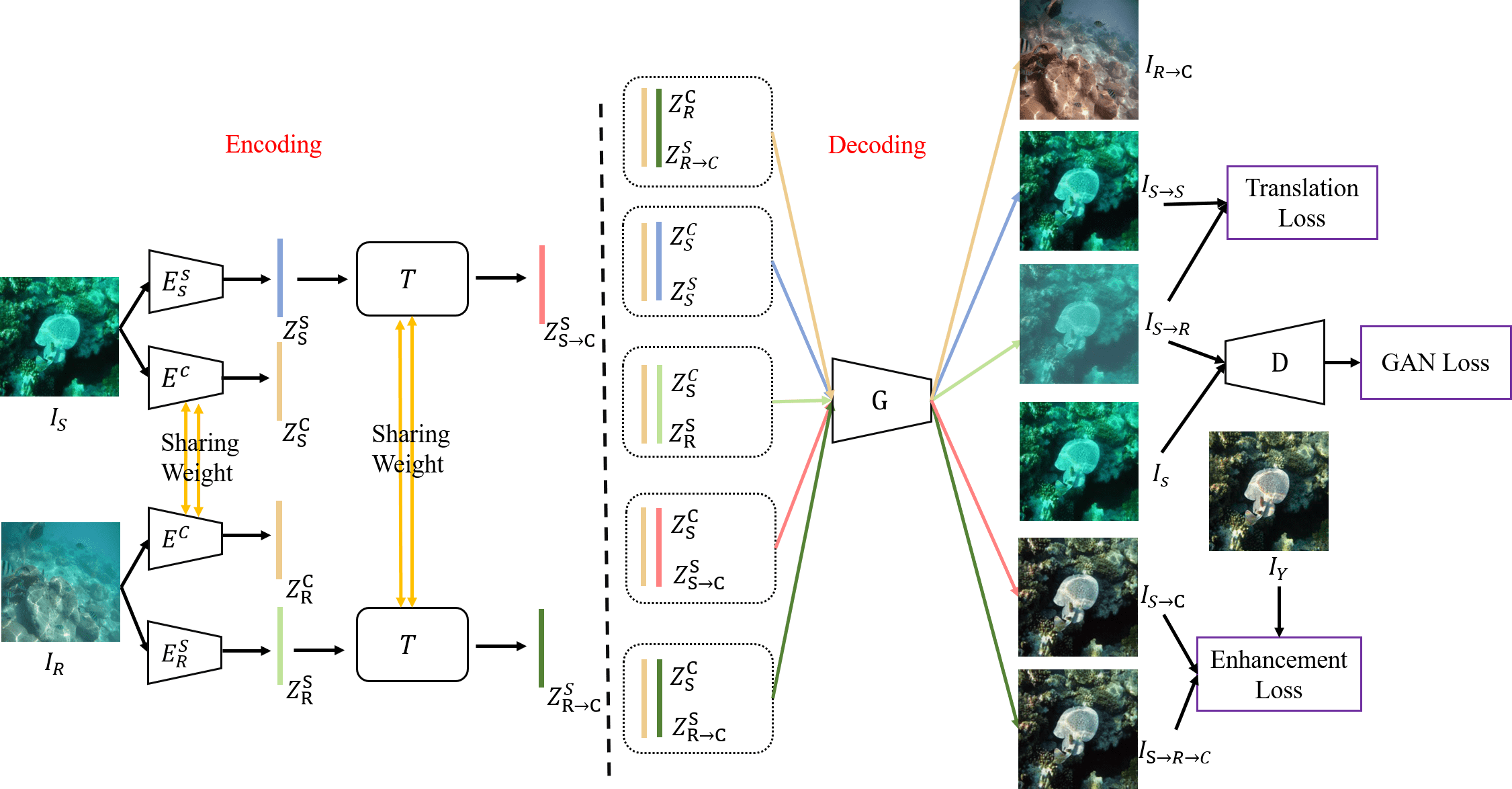}
  \captionsetup{justification=raggedright,singlelinecheck=true}
  \caption{The overview of our proposed framework. The framework consist two style encoder $E^{\scaleto{S}{4pt}}_{\scaleto{S}{4pt}}$ and $E^{\scaleto{S}{4pt}}_{\scaleto{R}{4pt}}$ and a share content encoder $E^{\scaleto{C}{4pt}}$, a latent transform unit $T$, a generator $G$ and a discriminator $D$. Given synthesis underwater image $I_{\scaleto{S}{4pt}}$ and real-world underwater image $I_{\scaleto{R}{4pt}}$, encoders first encode images into content and style latent space, and $T$ transform degraded latent to clean one. While decoding stage, it can input content latent and arbitrary style latent to the generator and output corresponding images. Image enhancement can be achieved by input content and clean style into generator and obtain real-world enhanced image $I_{\scaleto{R\rightarrow C}{4pt}}$ and synthesis enhanced image $I_{\scaleto{S\rightarrow C}{4pt}}$. We use different color to highlight different style latent.}
  \label{fig_model}
\end{figure*}

\section{Related Work}
\subsection{Underwater Image Enhancement}
\noindent Underwater image enhancement can roughly divide into three categories, which are non-physical model based, physical model based and learning based methods. Non-physical model based methods enhance pixel intensity directly to achieve better image quality without physical model constraint. \citet{fusion} proposed a fusion based method, which apply multi-scale fusion strategy on color corrected and contrast enhanced images. To modified the work of \cite{GLC21}, \citet{GLC22} proposed a contrast enhanced method that shape to follow the Rayleigh distribution in RGB space. Another technical line utilize Retinex theorem to design algorithms, \citet{waternet33} convert color corrected image to CIELab color space and enhance the L layer by Retinex theorem. Physical model based treat underwater image enhancement as an inverse problem, various priors and physical underwater image formation models are proposed. The well-known model is Jaffe-MaGlamery underwater image model \cite{GLC25, GLC26}, which can be expressed as
\begin{equation}
    I = Je^{-{\eta}d}+A(1-e^{-{\eta}d})
\end{equation}
where \textit{I} is the observed underwater image, \textit{J} is clean image, \textit{A} is back-scattered light, \textit{d} is distance between camera and the scene, and \textit{$\eta$} is light attenuation coefficient. \citet{seathru} revised the physical image formation model and proposed color correction method with RGBD images to restore degraded underwater image. Many image priors are also explored, \citet{GDCP} proposed a Generalized Dark Channel Prior (GDCP) and \citet{UDCP} proposed an Underwater Dark channel Prior (UDCP), which based on characteristic of wave-length dependent absorption that the signal of red channel is unreliable. 

Recent year deep learning based method demonstrate significant advancement in various computer vision tasks. The class of methods need large scale degraded and high-quality counterpart image pairs for training model, which is infeasible in real-world. To generate synthetic underwater image, \citet{waterGAN} proposed WaterGAN for synthesizing underwater image from in-air RGB-D image, the generator synthesize realistic underwater image by modeling light attenuation, light back scattering, and camera model, the similar method is adopted in \cite{UWGAN}. Rather than \cite{waterGAN, UWGAN}, \citet{UGAN} and \citet{FUnIE-GAN} utilize CycleGAN to learn the function of $f: I^{\scaleto{D}{4pt}}\rightarrow I^{\scaleto{C}{4pt}}$ to obtain synthesis underwater image pairs, where $I^{\scaleto{D}{4pt}}$ is in-air domain and $I^{\scaleto{C}{4pt}}$ is underwater domain. For underwater image enhancement, \citet{waterGAN} proposed a two-stage image restoration network, which include depth estimation and color restoration. \citet{waternet} use CNN to predict the confidence map of gamma corrected, histogram equalization and white balanced corrected image for fusion. \citet{deepRestoration_SingalLetter2019} proposed a channel-wise feature extraction module and utilize dense-residual block to enhance latent extracted under encoder-decoder architecture. \citet{dugan} proposed GAN-based model with a dual discriminator to address different level degradation caused by depth, and proposed a content loss and style loss for guiding training. Noting that the content and style loss \cite{dugan} is different from our content style disentangling setting, we focus on separate the image into different latent space for structural and appearance latent, i.e. content and style, and further perform domain translation and image enhancement.

Some works view underwater image enhancement in image-to-image translation perspective, the goal is building the mapping between underwater image domain and clean in-air domain. \citet{UGAN} utilize Generative Adversarial Network (GAN) for enhancing underwater image, which generator is a fully convolution U-Net alike encoder-decoder architecture, a similar work also done by \cite{UWGAN}. \citet{FUnIE-GAN} designed generator network by following principles of U-Net but employ fewer parameter and more efficient way to achieve fast inference. However, most of these works use synthesis image pairs for training, and ignore the domain gap between synthesis and real-world images, which make the performance degraded while testing on real-world data.

\subsection{Domain Adaptation for Underwater Image}
 \noindent To alleviate the domain shift problem between underwater and clean in-air image, many works are proposed to handle underwater image object recognition and detection \cite{detection0, detection2}. For underwater image enhancement, \citet{UIE-DAL} proposed a water-type auxiliary classifier to classify the encoded latent water-type is real or synthesis, the encoder would struggle to confuse classifier and be forced to generate domain-agnostic latent. \citet{physical} proposed a perceptual loss to minimize the discrepancy of encoded latent from synthesis and real domain, and use physical model feedback to constraint enhanced image should be closed to original input after transforming back to underwater domain via physical model. \citet{TUDA} observed that in addition to inter-domain gap between synthesis and real image, the complex real-world underwater environment also cause intra-domain gap, they proposed a two-phase underwater domain adaptation network with a rank-based underwater quality assessment method to solve the problem. Noting that the goal of all of these works is to learn a domain-agnostic latent, in other words, to minimize the discrepancy of latent encoded from synthesis and real domains, in contrast, we aim to separate image to content and style latent and distinguish style latent from different domains. \citet{two_step} proposed a two-step domain adaptation framework without synthesis data, they use CycleGAN for style transfer to remove color cast, which is similar with synthesizing underwater image (clean $\rightarrow$ underwater) in \cite{FUnIE-GAN} but in an opposite direction (underwater $\rightarrow$ clean), and remove hazy effect in second step.

\subsection{Style Separation}
\noindent Since recent years on the success of style transfer, it is shown that the content and style of image are separable. Learning the independent representation of content and style from input image is so-called style separation \cite{S3GAN} or content-style disentangling. Although it is difficult to define content and style, most of prior works \cite{MUNIT, MoCoGAN, DSN} assume content latent represent spatial structure of image, which is domain invariant; style latent represent the appearance of structure, which would be domain variant. To process spatial equivalent task, such as Image-to-Image translation (I2I), the goal is finding target domain style latent and decoding with original content latent. For style separation, Instance Normalization (IN) \cite{IN} is widely used for encoder, since the ability to wash out contrast-related information. In other hand, Adaptive Instance Normalization (AdaIN) \cite{AdaIN} is commonly used for decoder to reconstruct image for input content and style latent, because the strong ability to preserve style information. In this paper, we perform style separation on the real-world and synthesis underwater image to obtain content and style latent, and learn to build the mapping between underwater degraded style latent domain and in-air clean one, we adopt IN and AdaIN in our encoder and generator, the detail of our proposed framework would be elaborate in next section.

\section{Proposed Method}
\label{section_method}
\noindent As mentioned above, previous works \cite{FUnIE-GAN, UGAN, UWGAN} usually utilize synthetic pair data for training due to difficulty of collecting real-world pair data. Though some domain adaptation for underwater image enhancement model has been proposed \cite{UIE-DAL, physical, TUDA}, all of them aim to map synthesis and real domain images into one share latent space, and minimize discrepancy of latent encoded from these two domains, while under this manner, the encoded latent is lack of interpretability since the latent include domain variant and domain invariant information. In contrast to prior works, we argue that disentangle image into content and style \cite{MUNIT} is more effective representation. We build the model based on two assumption: 1) The image can be disentangled into content and style latent, and 2) images in different domains share same content latent space but with different domains in style latent space. For the assumption 1, many prior works \cite{MUNIT, S3GAN, MoCoGAN} has been empirically proved, and we will evaluate assumption 2 by visualizing style latent space. The proposed framework is illustrated in Fig ~\ref{fig_model}, based on the assumptions for content and style latent, the overall framework consist one share content encoder $E^{\scaleto{C}{4pt}}$ since the content latent space is domain invariant and shared for cross domains in our assumption, two style encoder $E_{\scaleto{S}{4pt}}^{\scaleto{S}{4pt}}$ and $E_{\scaleto{R}{4pt}}^{\scaleto{S}{4pt}}$, which encode image to style latent from synthesis or real-world underwater domain. Given synthesis underwater images $I_S$ and real-world underwater image $I_R$, we can obtain content latent $Z^{\scaleto{C}{4pt}}_{\scaleto{S}{4pt}}$, $Z^{\scaleto{C}{4pt}}_{\scaleto{R}{4pt}}$, and style latent $Z^{\scaleto{S}{4pt}}_{\scaleto{S}{4pt}}$, $Z^{\scaleto{S}{4pt}}_{\scaleto{R}{4pt}}$ from synthesis and real-world domain, then we introduce the latent transform unit $T$ to transform underwater style to clean domain, noting that the $T$ is a non-linear many-to-one mapping that performing latent level enhancement, in this case, $T$ is a two-to-one mapping, that is, the synthesis and real-world underwater style to clean style transformation. Thank to the simplicity and flexibility of designing $T$, it can be easy to extend to three-to-one or N-to-one mapping, and toward multi-target domain adaptation and domain generalization. We denote $Z^{\scaleto{S}{4pt}}_{\scaleto{S \rightarrow C}{4pt}}$ and $Z^{\scaleto{S}{4pt}}_{\scaleto{R \rightarrow C}{4pt}}$ as synthesis underwater to clean style and real-world underwater to clean style. After encoding image into latents, the generator $G$ is responsible to reconstruct target image in accordance with input content and style latent. Finally, discriminator $D$ is utilized to apply adversarial loss and distinguish realism of generated image, the generator is enforced to play a zero-sum game with discriminator to generate realistic target image.  

\subsection{Domain Adaptation and Image Enhancement}
\noindent We formulate image processing of proposed framework. Given $I_S$, high-quality counterpart $I_Y$ and $I_R$, we first extract individual image latent:
\begin{equation}
    Z^{\scaleto{C}{4pt}}_{\scaleto{S}{4pt}} = E^{\scaleto{C}{4pt}}(I_{\scaleto{S}{4pt}}),  Z^{\scaleto{C}{4pt}}_{\scaleto{R}{4pt}} = E^{\scaleto{C}{4pt}}(I_{\scaleto{R}{4pt}})
\end{equation}
\begin{equation}
    Z^{\scaleto{S}{4pt}}_{\scaleto{S}{4pt}} = E^{\scaleto{S}{4pt}}_{\scaleto{S}{4pt}}(I_{\scaleto{S}{4pt}}),  Z^{\scaleto{S}{4pt}}_{\scaleto{R}{4pt}} = E^{\scaleto{S}{4pt}}_{\scaleto{R}{4pt}}(I_{\scaleto{R}{4pt}})
\end{equation}
The content encode structural information, i.e. domain invariant, and the style encode appearance information, i.e. domain variant. Based on the assumption of the synthesis underwater, real-world underwater and clean style latent belongs to three different domains, there are two goals. First, learn the non-linear mapping $T$ between underwater and clean style, which can formula as:
\begin{equation}
    Z^{\scaleto{S}{4pt}}_ {{\varepsilon\rightarrow \scaleto{C}{4pt}}} = T(Z^{\scaleto{S}{4pt}}_{\varepsilon}), \varepsilon=\{S, R\}
\end{equation}

where $T$ is the non-linear latent transform function we want to learn. Second, training a generator which can distinguish style latent from different domains and generate image in accordance with input content and style latent. The intuition behind our design is that if generator can distinguish style latent from different domains, we can observe the relation between style latent from different domains, and interact with style by latent manipulation, which make the encoded latent more interpretable.

To perform domain adaptation, we can obtain pseudo real underwater image pair \{$I_{\scaleto{S \rightarrow R}{4pt}}$, $I_{\scaleto{Y}{4pt}}$\} to solve the problem of lacking ground truth of real-world image, and simply perform supervised learning to learn the real-world style enhancement, where $I_{\scaleto{S \rightarrow R}{4pt}}$ is syn2real underwater image and can be generated by:
\begin{equation}
    I_{\scaleto{S \rightarrow R}{4pt}} = G(Z^{\scaleto{C}{4pt}}_{\scaleto{S}{4pt}}, Z^{\scaleto{S}{4pt}}_{\scaleto{R}{4pt}})
\end{equation}

Noting that though the notion of performing domain adaptation via pseudo real-world image pair is similar with \cite{DAdehazing, two_step}, we highlight two difference from \cite{DAdehazing, two_step}. First, the input of enhancement module is different, the prior works input the domain translation result, i.e. syn2real image, into enhancement module and perform enhancement in image space, i.e. pixel level, in contrast, the input of our framework is style latent, which perform image enhancement in latent space, i.e. feature level. We argue that the feature level enhancement can be more easily and effective since the feature can be more representative than signal in image space. Second, prior works perform dual direction domain translation by training two translation networks and without latent space assumption, on the contrary, we work based on latent space assumption and perform domain translation and even image enhancement in a unified model. To perform image enhancement, it can be achieve through input content and transform enhanced style latent into generator.
\begin{comment}
\begin{equation}
    I_{\scaleto{S \rightarrow C}{4pt}} = G(Z^{\scaleto{C}{4pt}}_{\scaleto{S}{4pt}}, Z^{\scaleto{S}{4pt}}_{\scaleto{S \rightarrow C}{4pt}})
\end{equation}
\begin{equation}
    I_{\scaleto{R \rightarrow C}{4pt}} = G(Z^{\scaleto{C}{4pt}}_{\scaleto{R}{4pt}}, Z^{\scaleto{S}{4pt}}_{\scaleto{R \rightarrow C}{4pt}})
\end{equation}
\end{comment}

\subsection{Architecture} 
\subsubsection{Encoder}
\noindent We employ encoder proposed by \citet{MUNIT} to be content and style encoder. The content encoder consist numbers of residual blocks, each residual block is stacked by Conv-IN-ReLU, that Instance Normalization (IN) \cite{IN} has been discovered the property to wash out style information \cite{spade}. The style encoder do not utilize IN layer since the IN would wash out useful style information. The style encoder consist several Conv-ReLU blocks, and finally with average pooling layer down-sampling to target fixed output $d$-dimension vector.    
\subsubsection{Latent Transform Unit}
\noindent Latent transform unit is responsible to transform degraded style to clean style, where style latent $Z \in R^d$, that is, the latent transform unit can be expressed as $\Gamma: R^d \rightarrow R^d$, where $d$ is dimension of latent vector, we set $d$ to 8 in our experiment. Since the function input and output is related low dimension vector, we simply use MLP with skip connection as latent transform unit, which enjoy global feature extraction without burden parameters.       
\subsubsection{Generator}
\noindent The decoder proposed by \citet{MUNIT} is adopted as our generator, which contain a set of residual block, and a MLP. The residual blocks are formed by Up-sampling, ConV, AdaIN, and ReLU layer. Rather than IN, Adaptive Instance Normalization (AdaIN) \cite{AdaIN} is proposed to preserved and restore style, which can be represent as:
\begin{equation}
    AdaIN(x, \gamma, \beta) = \gamma (\frac{x-\mu(x)}{\sigma(x)}) + \beta
\end{equation}
where $x$ is input feature, $\mu$ and $\sigma$ is mean and standard variation,  $\gamma$ and $\beta$ is parameter dynamic generated by MLP and input style latent.
\subsubsection{Discriminator}
\noindent We use multi-scale discriminator proposed by \citet{discriminator} for discriminator $D$ and apply LSGAN objective \cite{LSGAN} as adversarial loss.   

\subsection{Training Losses}
The loss function contain two categories, which are image-to-image translation loss and image enhancement loss.
\subsubsection{Image-to-image Translation Loss}
\noindent This part loss functions are utilized for learning difference between synthesis and real-world underwater style latent, and image reconstruction, in other words, provide guidance for cross domain image-to-image translation.

\paragraph{Reconstruction Loss}
\noindent We impose self-reconstruction and cycle reconstruction loss to encourage model perform domain translation and generate corresponding output by input specific latent, the reconstruction loss can be formulated as two term:
\begin{equation}
    \begin{split}
        &\mathcal{L}_{cyc} = \mathbb{E}_{x \sim I_{\scaleto{S}{3pt}}, y \sim I_{\scaleto{R}{3pt}}}[\|I_{x \rightarrow y \rightarrow x} - x\|_1]  
        \\  &+ \mathbb{E}_{x \sim I_{\scaleto{S}{3pt}}, y \sim I_{\scaleto{R}{3pt}}}[\|I_{y \rightarrow x \rightarrow y} - y\|_1]
    \end{split}
\end{equation}
\begin{equation}
    \begin{split}
        &\mathcal{L}_{self}  = \mathbb{E}_{x \sim I_S}[\|I_{x\rightarrow x}- x\|_1] 
         + \mathbb{E}_{y \sim I_R}[\|I_{y\rightarrow y}- y\|_1]
    \end{split}
\end{equation}

\paragraph{Adversarial Loss}
\noindent We employ discriminator to evaluate image is real or generated, and make generator generate more realistic image. the adversarial loss can be imposed under adversarial training, which can be expressed as:
\begin{equation}
    \begin{split}
        &\mathcal{L}_{GAN} = \mathbb{E}_{x \sim I_{\scaleto{S}{3pt}},y \sim I_{\scaleto{R}{3pt}} }[log(1-D(I_{x\rightarrow y})] + \mathbb{E}_{y \sim I_{\scaleto{R}{3pt}}}[log(D(y))]
        \\ & + \mathbb{E}_{x \sim I_{\scaleto{S}{3pt}},y \sim I_{\scaleto{R}{3pt}} }[log(1-D(I_{y\rightarrow x})] + \mathbb{E}_{x \sim I_{\scaleto{S}{3pt}}}[log(D(x))]
    \end{split}
\end{equation}

The full image-to-image translation loss can be defined as follow:
\begin{equation}
    \begin{split}
    &\mathcal{L}_{tran} =  \mathcal{L}_{GAN} + \lambda_{self}\mathcal{L}_{self} + \mathcal{L}_{cyc}
    \end{split}
\end{equation}
where $\lambda_{self}$ is trade-off weighting.

\subsubsection{Image Enhancement Loss}
\noindent Loss functions in this part are employed to learn image reconstruction and the difference of style latent from underwater and clean domains, in other words, provide guidance for training latent transform unit $T$ and image enhancement. 

\paragraph{Image Quality Loss}
\noindent For image enhancement task, prior works \cite{loss_ssim_L1_0, loss_ssim_L1_1} supposed jointly use pixel fidelity loss and structure loss, e.g. SSIM, can achieve better result than only use pixel fidelity loss or structure loss. We adopt L1 norm for pixel fidelity loss since L2 norm is widely studied that might cause blurry result, and use SSIM loss for structural loss:
\begin{equation}
    \begin{split}
        &\mathcal{L}_{pixel} = \| I_{\scaleto{S \rightarrow C}{4pt}} - I_{\scaleto{C}{4pt}}\|_1
         + \| I_{\scaleto{S\rightarrow R \rightarrow C}{4pt}} - I_{\scaleto{C}{4pt}}\|_1
    \end{split}
\end{equation}
\begin{equation}
    \begin{split}
        &\mathcal{L}_{ssim} = 1-SSIM(I_{\scaleto{S \rightarrow C}{4pt}}, I_{\scaleto{C}{4pt}})
         + 1-SSIM(I_{\scaleto{S\rightarrow R \rightarrow C}{4pt}}, I_{\scaleto{C}{4pt}})
    \end{split}
\end{equation}
\begin{equation}
    \mathcal{L}_{iq} = \mathcal{L}_{ssim} + \mathcal{L}_{pixel}
\end{equation}
\paragraph{Perceptual Loss}
 \noindent Perceptual loss are widely used to minimize high-level semantic between enhanced image and ground truth for image enhancement, we impose content constraint on enhanced image. The perceptual loss can be expressed as:
\begin{equation}
    \begin{split}
        &\mathcal{L}_{per} = \mathop{\sum}\limits_{j} \frac{1}{c_j h_h w_j}\| \Phi_j(I_{\scaleto{S \rightarrow C}{4pt}}) - \Phi_j(I_{\scaleto{C}{4pt}})\|^2_2
        \\ & + \mathop{\sum}\limits_{l \in L_c} \| \Phi_j(I_{\scaleto{S\rightarrow R \rightarrow C}{4pt}}) - \Phi_j(I_{\scaleto{C}{4pt}})\|^2_2
    \end{split}
\end{equation}
where $c_j h_h w_j$ is shape of $j^{th}$ feature map $\Phi_j$ of pre-trained perceptual network.

\paragraph{TV Loss}
\noindent Total variation loss acts as smoothness regularization in our design, which aims to minimize gradient of image and reduce noise in enhanced image, which can express as:
\begin{equation}
    \begin{split}
        &\mathcal{L}_{tv} = \sum\limits_{i=1}^H \sum\limits_{j=1}^W \sqrt{(I_{i,j}-I_{i+1, j})^2 + (I_{i,j}-I_{i, j+1})^2}
    \end{split}
\end{equation}

\paragraph{Latent Loss}
\noindent As our assumption, the clean latent should be same domain no matter it come from synthesis domain or real-world domain, we further give distance constraint to clean latent from different domains, which can express as:
\begin{equation}
    \begin{split}
        &\mathcal{L}_{latent} = \| Z^{\scaleto{S}{4pt}}_{\scaleto{S \rightarrow C}{4pt}} - Z^{\scaleto{S}{4pt}}_{\scaleto{R \rightarrow C}{4pt}}\|_1
    \end{split}
\end{equation}

The full image enhancement loss can be defined as follow:

\begin{equation}
    \begin{split}
    &\mathcal{L}_{en} = \lambda_{latent} \mathcal{L}_{latent} + \lambda_{tv} \mathcal{L}_{tv} + \lambda_{per}\mathcal{L}_{per} + \lambda_{iq}\mathcal{L}_{iq}
    \end{split}
\end{equation}
where $\lambda_{latent}$, $\lambda_{tv}$, $\lambda_{per}$, $\lambda_{iq}$ are weightings to balance the contribution for each term.

Overall loss function is defined as follow:
\begin{equation}
    \begin{split}
    &\mathcal{L} = \mathcal{L}_{tran} + \mathcal{L}_{en} 
    \end{split}
\end{equation}

% real qual
\begin{table*}[!t]
\captionsetup{justification=raggedright,singlelinecheck=true}
\caption{Quality evaluation on real-world underwater benchmarks with different algorithms, the top three methods are marked as \textcolor{red}{RED}, \textcolor{Green}{GREEN}, and \textcolor{blue}{BLUE}}
\begin{center}
\begin{tabular}{|c|c|c|c|c|c|c|c|c|}
\hline
\textbf{Methods}&\multicolumn{4}{|c|}{\textbf{UIQM}  $\uparrow$} &\multicolumn{4}{|c|}{\textbf{UCIQE}$\uparrow$} \\
\cline{2-5}
\cline{6-9}
 & \textbf{\textit{UIEB}}& \textbf{\textit{EUVP}}& \textbf{\textit{Sea-thru}} & \textbf{\textit{SQUID}}&
\textbf{\textit{UIEB}}& \textbf{\textit{EUVP}}& \textbf{\textit{Sea-thru}} & \textbf{\textit{SQUID}}\\
\hline

UIBLA& 2.2518& 1.7821 & 1.5209 & 0.6879 & \textcolor{Green}{0.5926} & \textcolor{blue}{0.5549} & 0.4998 & 0.4669  \\
WaterNet & 3.4342 & 3.0882 & \textcolor{blue}{4.9881} & \textcolor{blue}{3.5276} & \textcolor{blue}{0.5711} & \textcolor{Green}{0.5793} & 0.5337 & \textcolor{Green}{0.5601}\\
FUnIE-GAN  & 3.6862 & 3.0211 & 4.0368 & 1.7504 & 0.5459 & 0.5086 & 0.5157 & 0.4659 \\
UWGAN  & \textcolor{blue}{3.6919} & \textcolor{blue}{3.2897} & 4.3374 & 2.7191 & 0.5655 & 0.5262 & \textcolor{Green}{0.5537} & \textcolor{blue}{0.5518}\\
UGAN  & 3.5459 & 2.9669 & 4.1940 & 1.9155 & 0.5597 & 0.5272 & 0.4970 & 0.4636 \\
UIE-DAL & \textcolor{red}{4.7201} & \textcolor{red}{4.3566} & \textcolor{red}{6.1529} & \textcolor{red}{5.7931} & 0.5472 & 0.5254 & \textcolor{blue}{0.5385} & 0.5147 \\
Ours & \textcolor{Green}{3.9165} & \textcolor{Green}{3.7102} & \textcolor{Green}{5.2948} & \textcolor{Green}{3.8408} & \textcolor{red}{0.5950} & \textcolor{red}{0.5918} & \textcolor{red}{0.5575} & \textcolor{red}{0.5990} \\
\hline
\end{tabular}
\label{table_real_qual}
\end{center}
\end{table*}

% synthesis qual
\begin{table}[hptb!]
\captionsetup{justification=raggedright,singlelinecheck=true}
\caption{Quality evaluation on synthesis dataset with different algorithms, the top three methods are marked as \textcolor{red}{RED}, \textcolor{Green}{GREEN}, and \textcolor{blue}{BLUE}}
\begin{center}
\begin{tabular}{|c|c|c|}
\hline
\textbf{Methods} & \textbf{\textit{PSNR}} $\uparrow$ & \textbf{\textit{SSIM}} $\uparrow$\\
\hline
UIBLA& 15.591 & 0.626  \\
WaterNet & 20.119 & 0.777 \\
FUnIE-GAN & \textcolor{Green}{23.623} & \textcolor{blue}{0.791} \\
UWGAN & 20.614 & 0.770\\
UGAN & \textcolor{blue}{23.518} & \textcolor{Green}{0.815} \\
UIE-DAL & 14.982 & 0.647 \\
Ours & \textcolor{red}{23.906}&\textcolor{red}{0.824} \\
\hline
\end{tabular}
\label{table_syn_qual}`
\end{center}
\end{table}

\section{Experiment}
\label{section_exp}
\noindent In this section, we provide implementation detail and analyze experiment result in quantity and quality, we selected six off-the-shelf algorithms to compare, include one physical model based method, i.e. UIBLA \cite{UIBLA}, one CNN fusion based method, i.e. Water-Net \cite{waternet}, three GAN based methods, i.e. UWGAN\cite{UWGAN}, UGAN \cite{UGAN} and FUnIE-GAN \cite{FUnIE-GAN}, and one domain adaptation for underwater image enhancement model, i.e. UIE-DAL \cite{UIE-DAL}. We use implementation of \cite{UIBLA_code} for UIBLA, and implementation of \cite{FUnIE-GAN} for UGAN, the others compared methods use the public training code with author recommend hyper-parameters or pre-trained model.    

\subsection{Implementation Details}
\noindent For training the model, we random crop the image with patch size 128$\times$128 and normalized to [0, 1]. We use Adam optimizer with betas 0.5 and 0.999, and set learning rate to 5e-4 and batch size to 1. Our implementation use Pytorch framework and the code is based on \cite{code}, all experiments are conducted on one NVIDIA GTX 2080Ti GPU and Intel i7-8700 CPU.   

We use the EUVP pair dataset \cite{FUnIE-GAN} as synthesis underwater dataset. EUVP dataset consist real-world unpair and synthesis image pair, there are 3.7K image pairs in underwater imagenet subset that is generated by CycleGAN, which synthesize real underwater appearance on good quality underwater image, the advantage of this dataset is that the most of images contain underwater scene or marine creature, which is more closed to real case, however, this synthesis dataset only contain color cast degradation, we further include part of data in synthesis dataset proposed by \citet{NYU_syn} using NYU-v2 RGB-D dataset to training set, this dataset contain multiple degradation, different color cast and different level hazy effect. As for real-world dataset, we use Underwater Image Enhancement Benchmark dataset (UIEB) \cite{waternet}, which consist 950 unpair real-world underwater images.

\subsection{Quality Comparison}
\noindent For synthesis underwater image quality assessment, we use PSNR and SSIM to evaluate, which are widely used full-reference metrics for image enhancement, and we adopt Underwater Image Quality Measures (UIQM) and Underwater Color Image Quality Evaluation (UCIQE) to evaluate real underwater image, which are widely used no-reference metrics for underwater image enhancement.

% real eval.
we evaluate proposed method on four public real-world underwater benchmarks, i.e. UIEB \cite{waternet}, EUVP \cite{FUnIE-GAN}, SQUID \cite{SQUID}, Sea-thru \cite{seathru}. Table \ref{table_real_qual} show the evaluation result on real-world datasets. As the table shown, our method outperform all other methods for UCIQE and with large margin on SQUID dataset, and achieve second place for UIQM. Despite the fact that UIE-DAL result high UIQM score, it loss structure of original image and do not match to human visual system. We suppose the reason is that UIQM is linear combination of UICM, UISM and UIConM, which scored by colorfulness, sharpness and contrast, the generated high saturation image can achieve high score in these three terms but not guarantee to achieve high naturalness or maintain structural information, which might be fail the human visual evaluation. For UCIQE, our proposed method outperform all of other comparison algorithms for all datasets, especial for SQUID dataset, since the dataset is composed by heavily degraded underwater image, and most of others methods fail to enhanced.    

% synthesis eval
Table \ref{table_syn_qual} show the evaluation of synthesis pair data from EUVP underwater imagenet set, our method outperform all of comparison method, and be competitive with FUnIE-GAN and UGAN, however, these two model over-fit on synthetic data and hard to generalize to real-world underwater image, which performance is much worse than our and shown in Table \ref{table_real_qual}, the limitation block the general usage of real-world application for this two methods. UIE-DAL fail to enhance synthetic underwater image since lack of pixel fidelity to original image and loss structural information. Our method not only perform well on real-world image but also synthetic dataset, and demonstrate the success of bridging the domain gap between synthetic and real-world data.

%SYN visualization
\begin{figure*}[!t]
    \begin{subfigure}[b]{0.105\textwidth}
        \centering
        \begin{minipage}{1.0\textwidth}
          \vspace*{\fill}
          \centering
          \includegraphics[width=1.0\linewidth]{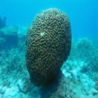} 
          \par\vspace{1mm}
          \includegraphics[width=1.0\linewidth]{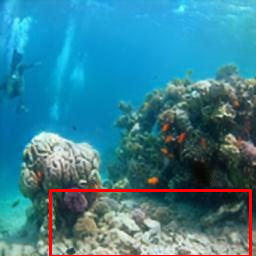}
          \par\vspace{1mm}
          \includegraphics[width=1.0\linewidth]{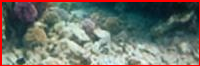}
        \end{minipage}
        \captionsetup{labelformat=empty}
        \caption{Synthesis underwater image}
    \vspace{-0.63cm}
    \end{subfigure}
    \hfill
    \begin{subfigure}[b]{0.105\textwidth}
        \centering
        \begin{minipage}{1.0\textwidth}
              \vspace*{\fill}
              \centering
              \includegraphics[width=1.0\linewidth]{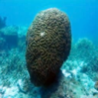} 
              \par\vspace{1mm}
              \includegraphics[width=1.0\linewidth]{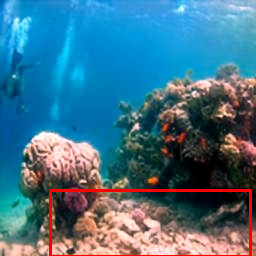}
              \par\vspace{1mm}
              \includegraphics[width=1.0\linewidth]{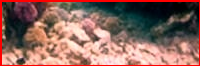}
            \end{minipage}
        \captionsetup{labelformat=empty}
        \caption{UIBLA \cite{UIBLA}}
    \end{subfigure}
    \hfill
    \begin{subfigure}[b]{0.105\textwidth}
        \centering
        \begin{minipage}{1.0\textwidth}
          \vspace*{\fill}
          \centering
          \includegraphics[width=1.0\linewidth]{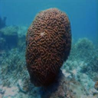} 
          \par\vspace{1mm}
          \includegraphics[width=1.0\linewidth]{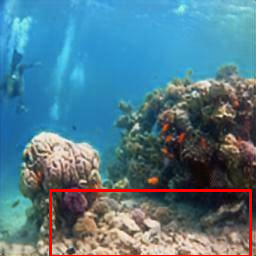}
          \par\vspace{1mm}
          \includegraphics[width=1.0\linewidth]{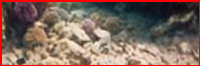}
        \end{minipage}
        \captionsetup{labelformat=empty}
        \caption{Water-Net \cite{waternet}}
    \end{subfigure}
    \hfill
    \begin{subfigure}[b]{0.105\textwidth}
        \centering
        \begin{minipage}{1.0\textwidth}
              \vspace*{\fill}
              \centering
              \includegraphics[width=1.0\linewidth]{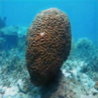} 
              \par\vspace{1mm}
              \includegraphics[width=1.0\linewidth]{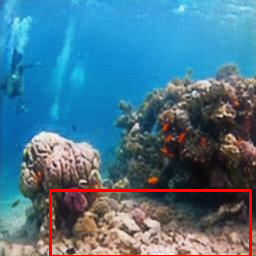} 
              \par\vspace{1mm}
              \includegraphics[width=1.0\linewidth]{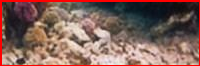} 
            \end{minipage} 
        \captionsetup{labelformat=empty}
        \caption{FUnIE-GAN \cite{FUnIE-GAN}}
    \end{subfigure}
    \hfill
    \begin{subfigure}[b]{0.105\textwidth}
        \centering
        \begin{minipage}{1.0\textwidth}
          \vspace*{\fill}
          \centering
          \includegraphics[width=1.0\linewidth]{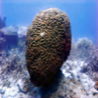} 
          \par\vspace{1mm}
          \includegraphics[width=1.0\linewidth]{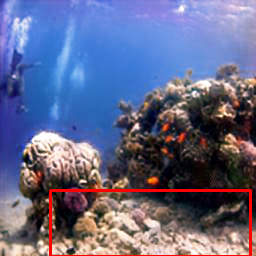}
          \par\vspace{1mm}
          \includegraphics[width=1.0\linewidth]{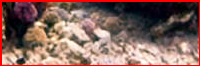}
        \end{minipage}
        \captionsetup{labelformat=empty}
        \caption{UWGAN \cite{UWGAN}}
    \end{subfigure}
    \hfill
    \begin{subfigure}[b]{0.105\textwidth}
        \centering
        \begin{minipage}{1.0\textwidth}
          \vspace*{\fill}
          \centering
          \includegraphics[width=1.0\linewidth]{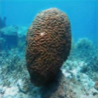} 
          \par\vspace{1mm}
          \includegraphics[width=1.0\linewidth]{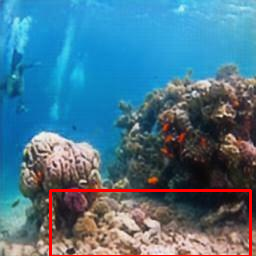}
          \par\vspace{1mm}
          \includegraphics[width=1.0\linewidth]{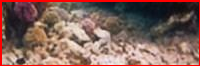}
        \end{minipage} 
        \captionsetup{labelformat=empty}
        \caption{UGAN \cite{UGAN}}
    \end{subfigure}
    \hfill
   \begin{subfigure}[b]{0.105\textwidth}
   \centering
        \begin{minipage}{1.0\textwidth}
          \vspace*{\fill}
          \centering
          \includegraphics[width=1.0\linewidth]{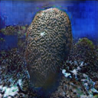} 
          \par\vspace{1mm}
          \includegraphics[width=1.0\linewidth]{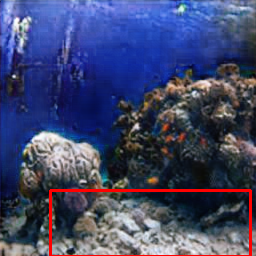}
          \par\vspace{1mm}
          \includegraphics[width=1.0\linewidth]{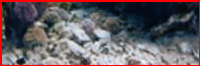}
        \end{minipage}
        \captionsetup{labelformat=empty}
        \caption{UIE-DAL \cite{UIE-DAL}}
    \end{subfigure}
    \hfill
    \begin{subfigure}[b]{0.105\textwidth}
        \centering
        \begin{minipage}{1.0\textwidth}
              \vspace*{\fill}
              \centering
              \includegraphics[width=1.0\linewidth]{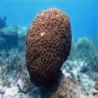} 
              \par\vspace{1mm}
              \includegraphics[width=1.0\linewidth]{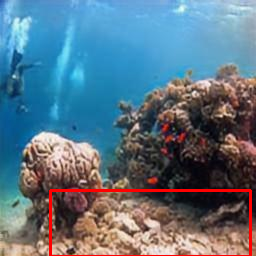}
              \par\vspace{1mm}
              \includegraphics[width=1.0\linewidth]{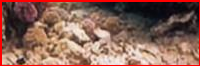}
            \end{minipage} 
        \captionsetup{labelformat=empty}
        \caption{UIESS (Ours)}
    \end{subfigure}
    \hfill
    \begin{subfigure}[b]{0.105\textwidth}
        \centering
        \begin{minipage}{1.0\textwidth}
          \vspace*{\fill}
          \centering
          \includegraphics[width=1.0\linewidth]{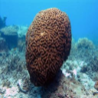} 
          \par\vspace{1mm}
          \includegraphics[width=1.0\linewidth]{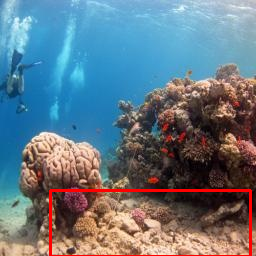}
          \par\vspace{1mm}
          \includegraphics[width=1.0\linewidth]{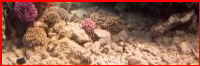}
        \end{minipage} 
        \captionsetup{labelformat=empty}
        \caption{GT}
    \end{subfigure}
  \caption{Visual comparisons of synthesis data.}
  \label{fig_syn_quan}
\end{figure*}

%real world: UIEB EUVP
\begin{figure*}[t!]
    \begin{subfigure}[b]{0.120\textwidth}
        \centering
        \begin{minipage}[b][3cm][b]{1.0\textwidth}
              \vspace*{\fill}
              \centering
              \includegraphics[width=1.0\linewidth]{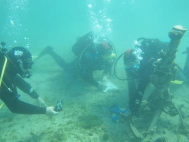} 
              \par\vspace{1mm}
              \includegraphics[width=1.0\linewidth]{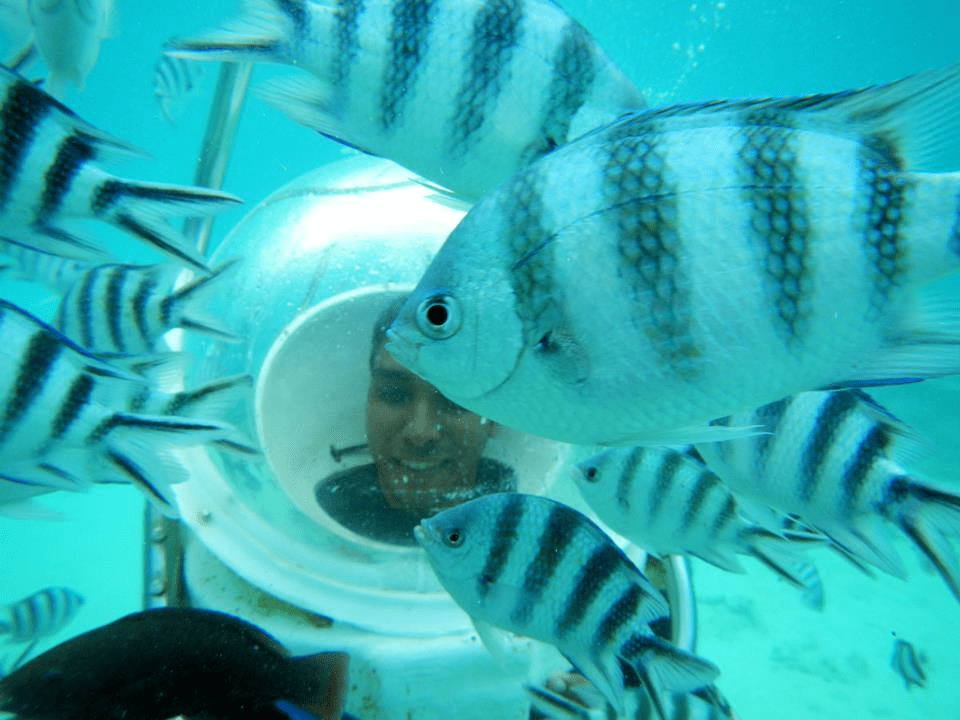} 
        \end{minipage}
        \captionsetup{labelformat=empty}
        \caption{Real underwater image}
        \vspace{-0.32cm}
    \end{subfigure}
    \hfill
    \hspace{-0.25cm}
    \begin{subfigure}[b]{0.120\textwidth}
        \centering
        \begin{minipage}[b][3cm][b]{1.0\textwidth}
          \vspace*{\fill}
          \centering
          \includegraphics[width=1.0\linewidth]{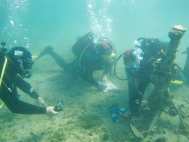} 
          \par\vspace{1mm}
          \includegraphics[width=1.0\linewidth]{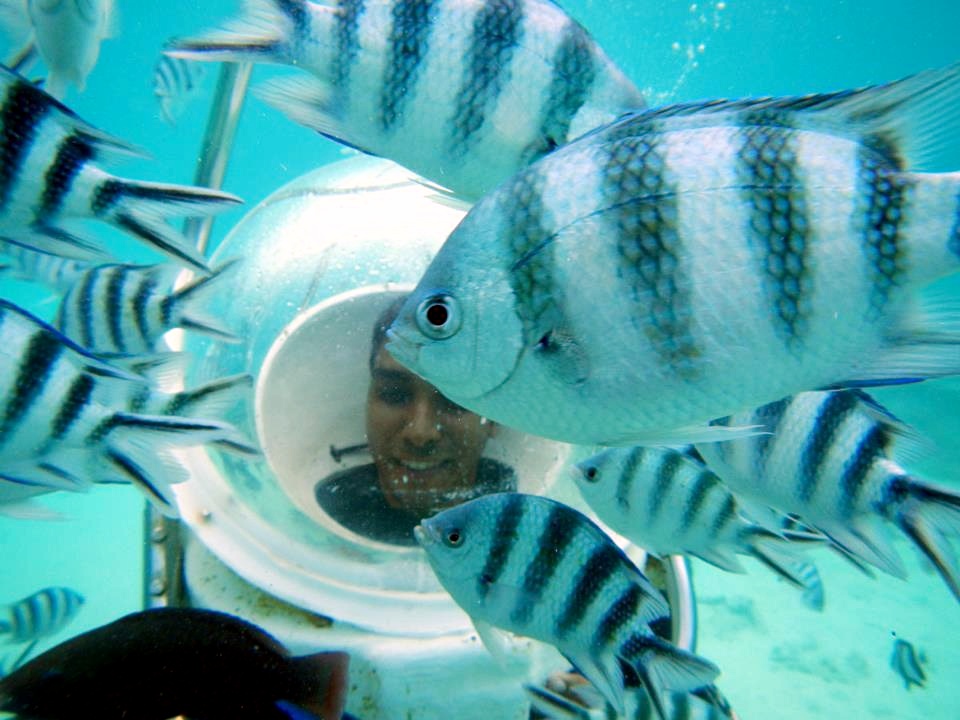} 
        \end{minipage}
        \captionsetup{labelformat=empty}
        \caption{UIBLA \cite{UIBLA}}
    \end{subfigure}
    \hfill
    \hspace{-0.25cm}
    \begin{subfigure}[b]{0.120\textwidth}
        \centering
        \begin{minipage}[b][3cm][b]{1.0\textwidth}
          \vspace*{\fill}
          \centering
          \includegraphics[width=1.0\linewidth]{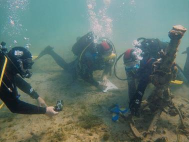} 
          \par\vspace{1mm}
          \includegraphics[width=1.0\linewidth]{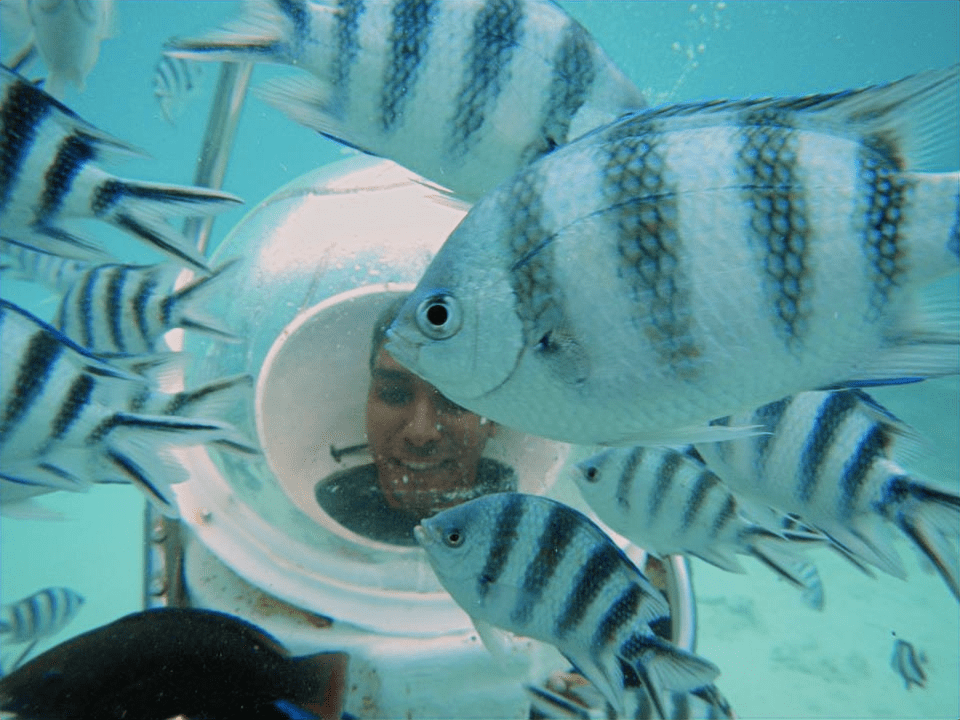} 
        \end{minipage} 
        \captionsetup{labelformat=empty}
        \caption{Water-Net \cite{waternet}}
    \end{subfigure}
    \hfill
    \hspace{-0.25cm}
    \begin{subfigure}[b]{0.120\textwidth}
        \centering
        \begin{minipage}[b][3cm][b]{1.0\textwidth}
          \vspace*{\fill}
          \centering
          \includegraphics[width=1.0\linewidth]{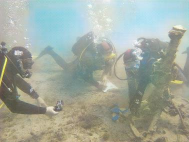} 
          \par\vspace{1mm}
          \includegraphics[width=1.0\linewidth]{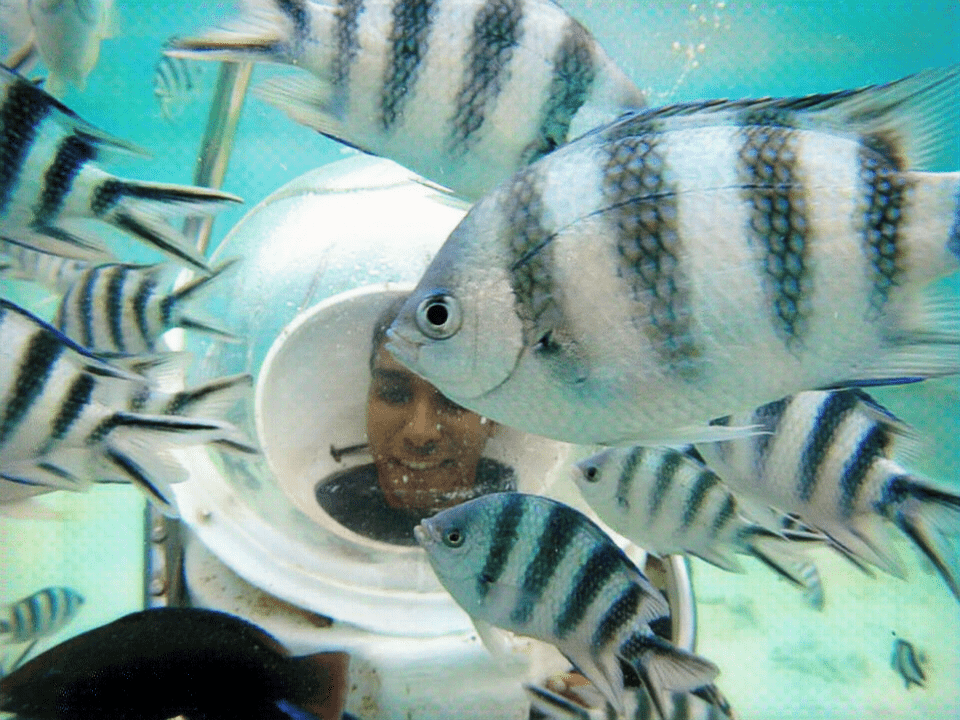} 
        \end{minipage} 
        \captionsetup{labelformat=empty}
        \caption{FUnIE-GAN \cite{FUnIE-GAN}}
    \end{subfigure}
    \hfill
    \hspace{-0.25cm}
    \begin{subfigure}[b]{0.120\textwidth}
        \centering
        \begin{minipage}[b][3cm][b]{1.0\textwidth}
          \vspace*{\fill}
          \centering
          \includegraphics[width=1.0\linewidth]{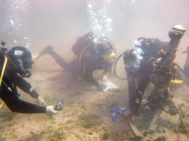} 
          \par\vspace{1mm}
          \includegraphics[width=1.0\linewidth]{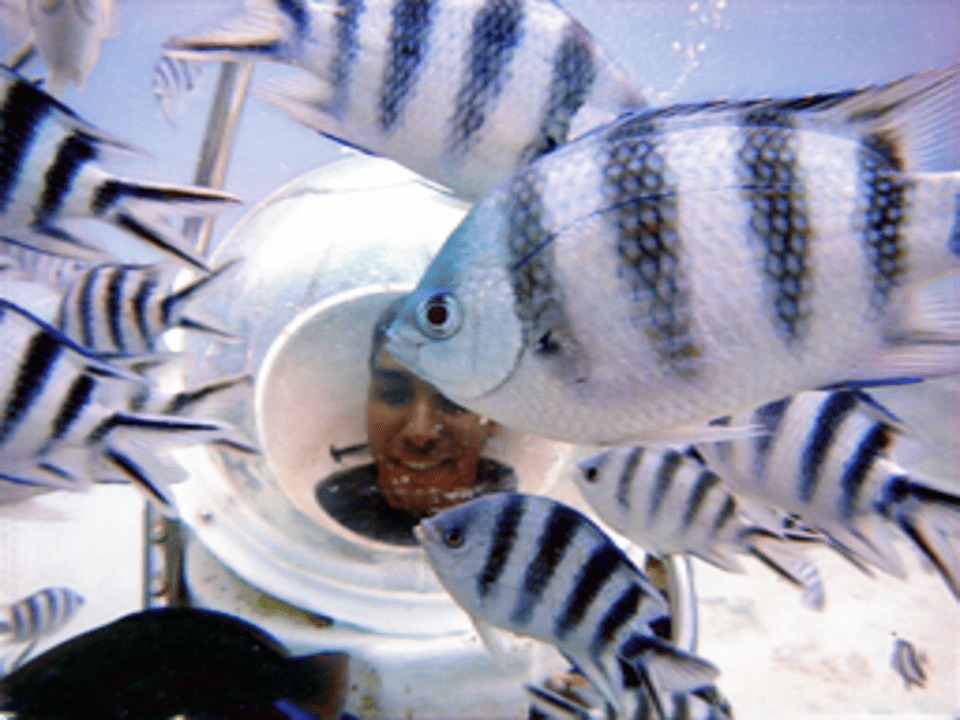} 
        \end{minipage} 
        \captionsetup{labelformat=empty}
        \caption{UWGAN \cite{UWGAN}}
    \end{subfigure}
    \hfill
    \hspace{-0.25cm}
    \begin{subfigure}[b]{0.120\textwidth}
        \centering
        \begin{minipage}[b][3cm][b]{1.0\textwidth}
          \vspace*{\fill}
          \centering
          \includegraphics[width=1.0\linewidth]{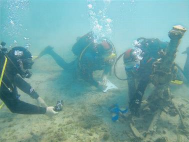} 
          \par\vspace{1mm}
          \includegraphics[width=1.0\linewidth]{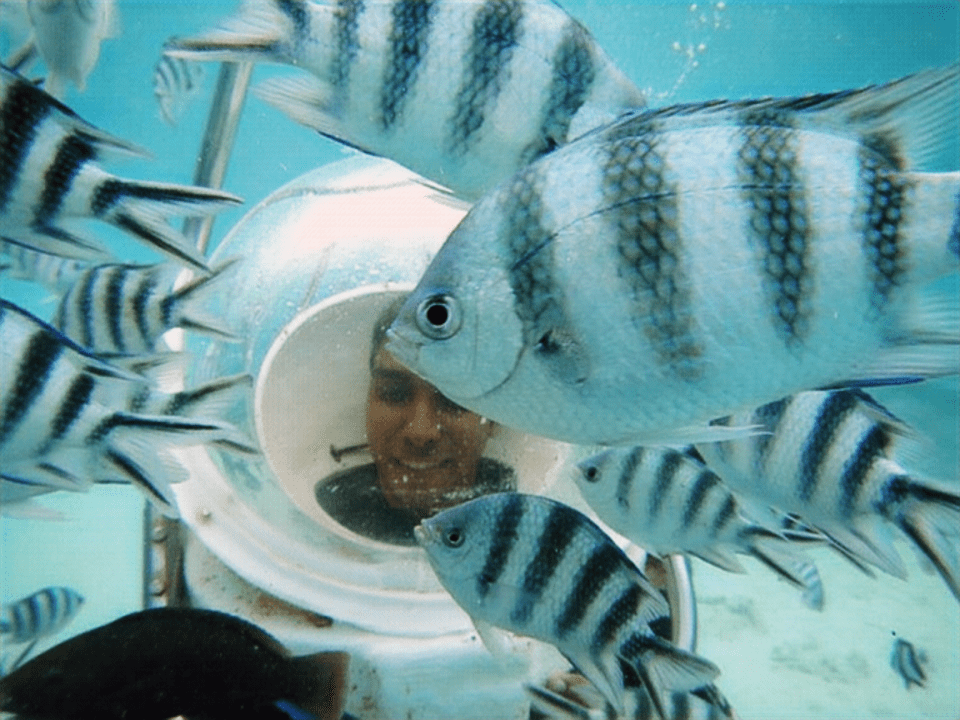} 
        \end{minipage} 
        \captionsetup{labelformat=empty}
        \caption{UGAN \cite{UGAN}}
    \end{subfigure}
    \hfill
    \hspace{-0.25cm}
   \begin{subfigure}[b]{0.120\textwidth}
       \centering
        \begin{minipage}[b][3cm][b]{1.0\textwidth}
          \vspace*{\fill}
          \centering
          \includegraphics[width=1.0\linewidth]{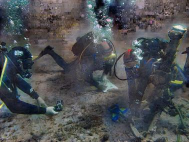} 
          \par\vspace{1mm}
          \includegraphics[width=1.0\linewidth]{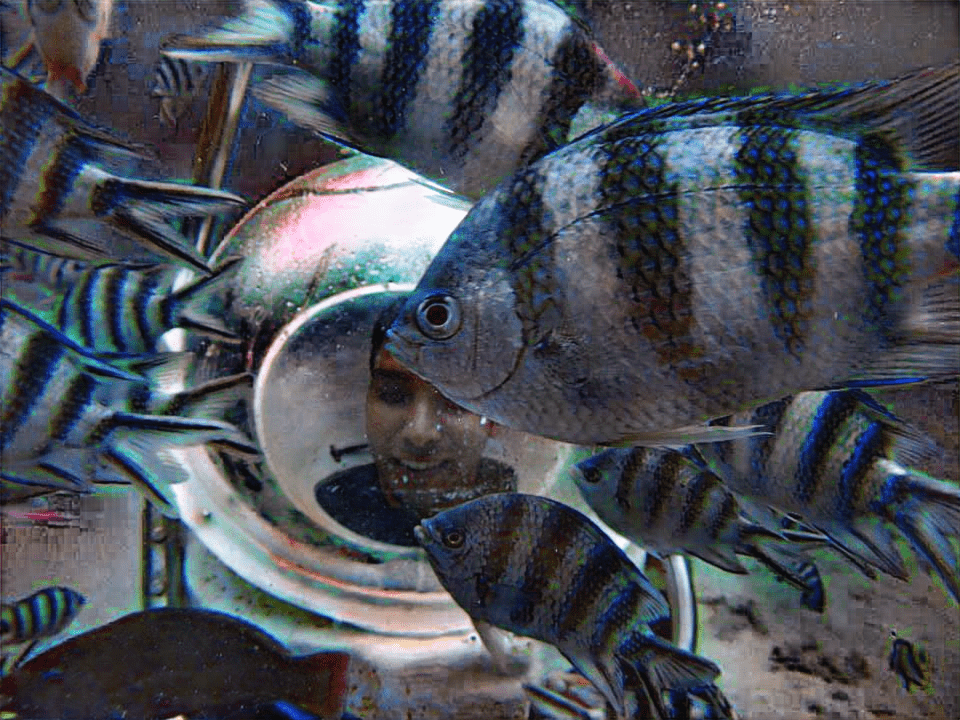} 
        \end{minipage} 
        \captionsetup{labelformat=empty}
        \caption{UIE-DAL \cite{UIE-DAL}}
    \end{subfigure}
    \hfill
    \hspace{-0.25cm}
    \begin{subfigure}[b]{0.120\textwidth}
        \centering
        \begin{minipage}[b][3cm][b]{1.0\textwidth}
          \vspace*{\fill}
          \centering
          \includegraphics[width=1.0\linewidth]{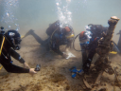} 
          \par\vspace{1mm}
          \includegraphics[width=1.0\linewidth]{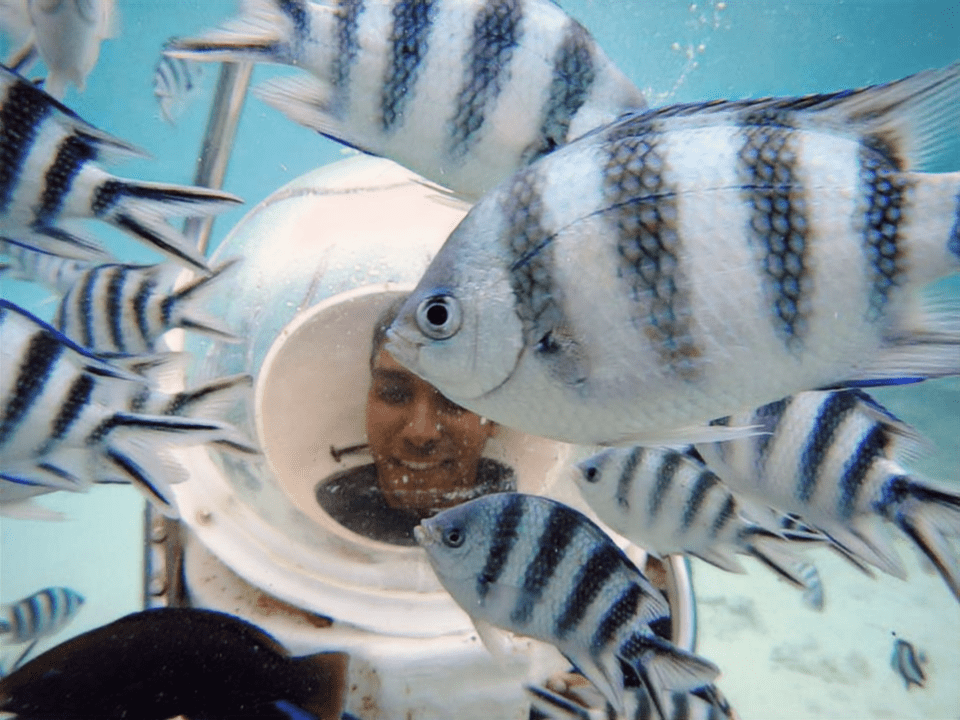} 
        \end{minipage}
        \captionsetup{labelformat=empty}
        \caption{UIESS (Ours)}
    \end{subfigure}
  \caption{Visual comparisons of real-world data from UIEB dataset.}
  \label{fig_real_quan_UIEB}
\end{figure*}

\begin{figure*}[t!]
    \begin{subfigure}[b]{0.120\textwidth}
        \centering
        \begin{minipage}[b][3cm][b]{1.0\textwidth}
          \vspace*{\fill}
          \centering
          \includegraphics[width=1.0\linewidth]{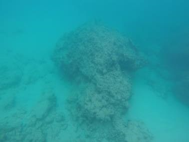} 
          \par\vspace{1mm}
          \includegraphics[width=1.0\linewidth]{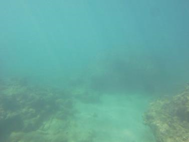} 
        \end{minipage}
        \captionsetup{labelformat=empty}
       \caption{Real underwater image}
       \vspace{-0.32cm}
    \end{subfigure}
    \hfill
    \hspace{-0.25cm}
    \begin{subfigure}[b]{0.120\textwidth}
        \centering
        \begin{minipage}[b][3cm][b]{1.0\textwidth}
          \vspace*{\fill}
          \centering
          \includegraphics[width=1.0\linewidth]{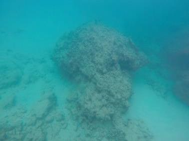} 
          \par\vspace{1mm}
          \includegraphics[width=1.0\linewidth]{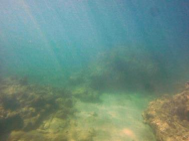} 
        \end{minipage} 
        \captionsetup{labelformat=empty}
        \caption{UIBLA \cite{UIBLA}}
    \end{subfigure}
    \hfill
    \hspace{-0.25cm}
    \begin{subfigure}[b]{0.120\textwidth}
        \centering
        \begin{minipage}[b][3cm][b]{1.0\textwidth}
          \vspace*{\fill}
          \centering
          \includegraphics[width=1.0\linewidth]{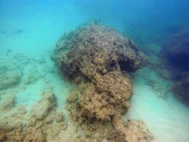} 
          \par\vspace{1mm}
          \includegraphics[width=1.0\linewidth]{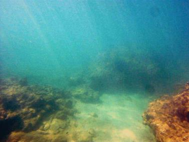} 
        \end{minipage} 
        \captionsetup{labelformat=empty}
        \caption{Water-Net \cite{waternet}}
    \end{subfigure}
    \hfill
    \hspace{-0.25cm}
    \begin{subfigure}[b]{0.120\textwidth}
        \centering
        \begin{minipage}[b][3cm][b]{1.0\textwidth}
          \vspace*{\fill}
          \centering
          \includegraphics[width=1.0\linewidth]{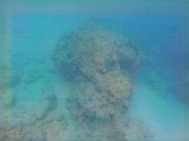} 
          \par\vspace{1mm}
          \includegraphics[width=1.0\linewidth]{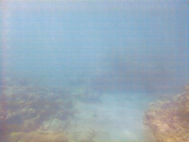} 
        \end{minipage}
        \captionsetup{labelformat=empty}
        \caption{FUnIE-GAN \cite{FUnIE-GAN}}
    \end{subfigure}
    \hfill
    \hspace{-0.25cm}
    \begin{subfigure}[b]{0.120\textwidth}
        \centering
        \begin{minipage}[b][3cm][b]{1.0\textwidth}
            \vspace*{\fill}
            \centering
            \includegraphics[width=1.0\linewidth]{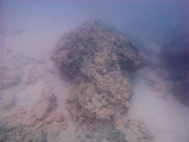} 
            \par\vspace{1mm}
            \includegraphics[width=1.0\linewidth]{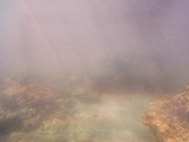} 
        \end{minipage}
        \captionsetup{labelformat=empty}
        \caption{UWGAN \cite{UWGAN}}
    \end{subfigure}
    \hfill
    \hspace{-0.25cm}
    \begin{subfigure}[b]{0.120\textwidth}
        \centering
        \begin{minipage}[b][3cm][b]{1.0\textwidth}
          \vspace*{\fill}
          \centering
          \includegraphics[width=1.0\linewidth]{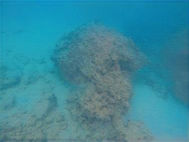} 
          \par\vspace{1mm}
          \includegraphics[width=1.0\linewidth]{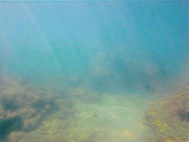} 
        \end{minipage}
        \captionsetup{labelformat=empty}
        \caption{UGAN \cite{UGAN}}
    \end{subfigure}
    \hfill
    \hspace{-0.25cm}
   \begin{subfigure}[b]{0.120\textwidth}
       \centering
        \begin{minipage}[b][3cm][b]{1.0\textwidth}
          \vspace*{\fill}
          \centering
          \includegraphics[width=1.0\linewidth]{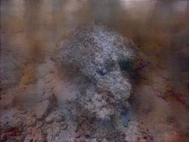} 
          \par\vspace{1mm}
          \includegraphics[width=1.0\linewidth]{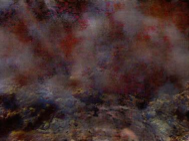} 
        \end{minipage}
        \captionsetup{labelformat=empty}
        \caption{UIE-DAL \cite{UIE-DAL}}
    \end{subfigure}
    \hfill
    \hspace{-0.25cm}
    \begin{subfigure}[b]{0.120\textwidth}
        \centering
        \begin{minipage}[b][3cm][b]{1.0\textwidth}
          \vspace*{\fill}
          \centering
          \includegraphics[width=1.0\linewidth]{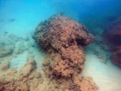} 
          \par\vspace{1mm}
          \includegraphics[width=1.0\linewidth]{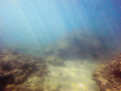} 
        \end{minipage}
        \captionsetup{labelformat=empty}
        \caption{UIESS (Ours)}
    \end{subfigure}
  \caption{Visual comparisons of real-world data from EUVP unpair dataset.}
  \label{fig_real_quan_EUVP}
\end{figure*}

%real world: sea_thru SQUID
\begin{figure*}[t!]
    \begin{subfigure}[b]{0.120\textwidth}
        \centering
        \begin{minipage}[b][3cm][b]{1.0\textwidth}
          \vspace*{\fill}
          \centering
          \includegraphics[width=1.0\linewidth]{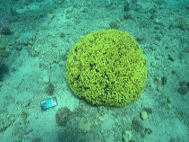} 
          \par\vspace{1mm}
          \includegraphics[width=1.0\linewidth]{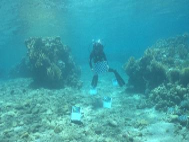} 
        \end{minipage}
        \captionsetup{labelformat=empty}
        \caption{Real underwater image}
        \vspace{-0.32cm}
    \end{subfigure}
    \hfill
    \hspace{-0.25cm}
    \begin{subfigure}[b]{0.120\textwidth}
        \centering
        \begin{minipage}[b][3cm][b]{1.0\textwidth}
          \vspace*{\fill}
          \centering
          \includegraphics[width=1.0\linewidth]{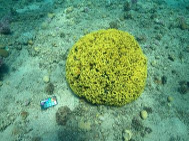} 
          \par\vspace{1mm}
          \includegraphics[width=1.0\linewidth]{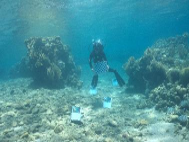} 
        \end{minipage} 
        \captionsetup{labelformat=empty}
        \caption{UIBLA \cite{UIBLA}}
    \end{subfigure}
    \hfill
    \hspace{-0.25cm}
    \begin{subfigure}[b]{0.120\textwidth}
        \centering
        \begin{minipage}[b][3cm][b]{1.0\textwidth}
          \vspace*{\fill}
          \centering
          \includegraphics[width=1.0\linewidth]{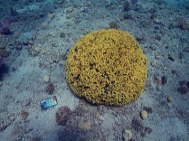} 
          \par\vspace{1mm}
          \includegraphics[width=1.0\linewidth]{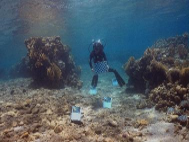} 
        \end{minipage}
        \captionsetup{labelformat=empty}
        \caption{Water-Net \cite{waternet}}
    \end{subfigure}
    \hfill
    \hspace{-0.25cm}
    \begin{subfigure}[b]{0.120\textwidth}
        \centering
        \begin{minipage}[b][3cm][b]{1.0\textwidth}
          \vspace*{\fill}
          \centering
          \includegraphics[width=1.0\linewidth]{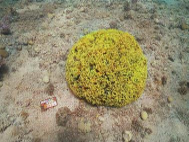} 
          \par\vspace{1mm}
          \includegraphics[width=1.0\linewidth]{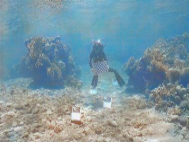} 
        \end{minipage}
        \captionsetup{labelformat=empty}
        \caption{FUnIE-GAN \cite{FUnIE-GAN}}
    \end{subfigure}
    \hfill
    \hspace{-0.25cm}
    \begin{subfigure}[b]{0.120\textwidth}
        \centering
        \begin{minipage}[b][3cm][b]{1.0\textwidth}
          \vspace*{\fill}
          \centering
          \includegraphics[width=1.0\linewidth]{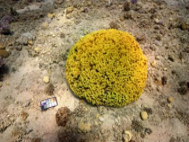} 
          \par\vspace{1mm}
          \includegraphics[width=1.0\linewidth]{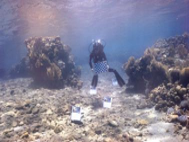} 
        \end{minipage}
        \captionsetup{labelformat=empty}
        \caption{UWGAN \cite{UWGAN}}
    \end{subfigure}
    \hfill
    \hspace{-0.25cm}
    \begin{subfigure}[b]{0.120\textwidth}
        \centering
        \begin{minipage}[b][3cm][b]{1.0\textwidth}
          \vspace*{\fill}
          \centering
          \includegraphics[width=1.0\linewidth]{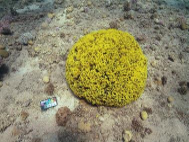} 
          \par\vspace{1mm}
          \includegraphics[width=1.0\linewidth]{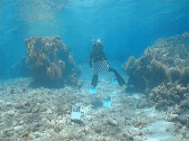} 
        \end{minipage}
        \captionsetup{labelformat=empty}
        \caption{UGAN \cite{UGAN}}
    \end{subfigure}
    \hfill
    \hspace{-0.25cm}
   \begin{subfigure}[b]{0.120\textwidth}
       \centering
        \begin{minipage}[b][3cm][b]{1.0\textwidth}
          \vspace*{\fill}
          \centering
          \includegraphics[width=1.0\linewidth]{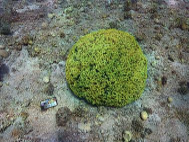} 
          \par\vspace{1mm}
          \includegraphics[width=1.0\linewidth]{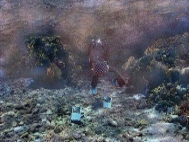} 
        \end{minipage}
        \captionsetup{labelformat=empty}
        \caption{UIE-DAL \cite{UIE-DAL}}
    \end{subfigure}
    \hfill
    \hspace{-0.25cm}
    \begin{subfigure}[b]{0.120\textwidth}
        \centering
        \begin{minipage}[b][3cm][b]{1.0\textwidth}
          \vspace*{\fill}
          \centering
          \includegraphics[width=1.0\linewidth]{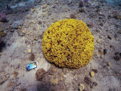} 
          \par\vspace{1mm}
          \includegraphics[width=1.0\linewidth]{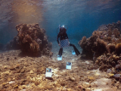} 
        \end{minipage}
        \captionsetup{labelformat=empty}
        \caption{UIESS (Ours)}
    \end{subfigure}
  \caption{Visual comparisons of real-world data from Sea-thru dataset.}
  \label{fig_real_quan_Seathru}
\end{figure*}

\begin{figure*}[t!]
    \begin{subfigure}[b]{0.120\textwidth}
        \centering
        \begin{minipage}[b][3cm][b]{1.0\textwidth}
          \vspace*{\fill}
          \centering
          \includegraphics[width=1.0\linewidth]{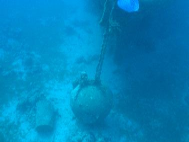} 
          \par\vspace{1mm}
          \includegraphics[width=1.0\linewidth]{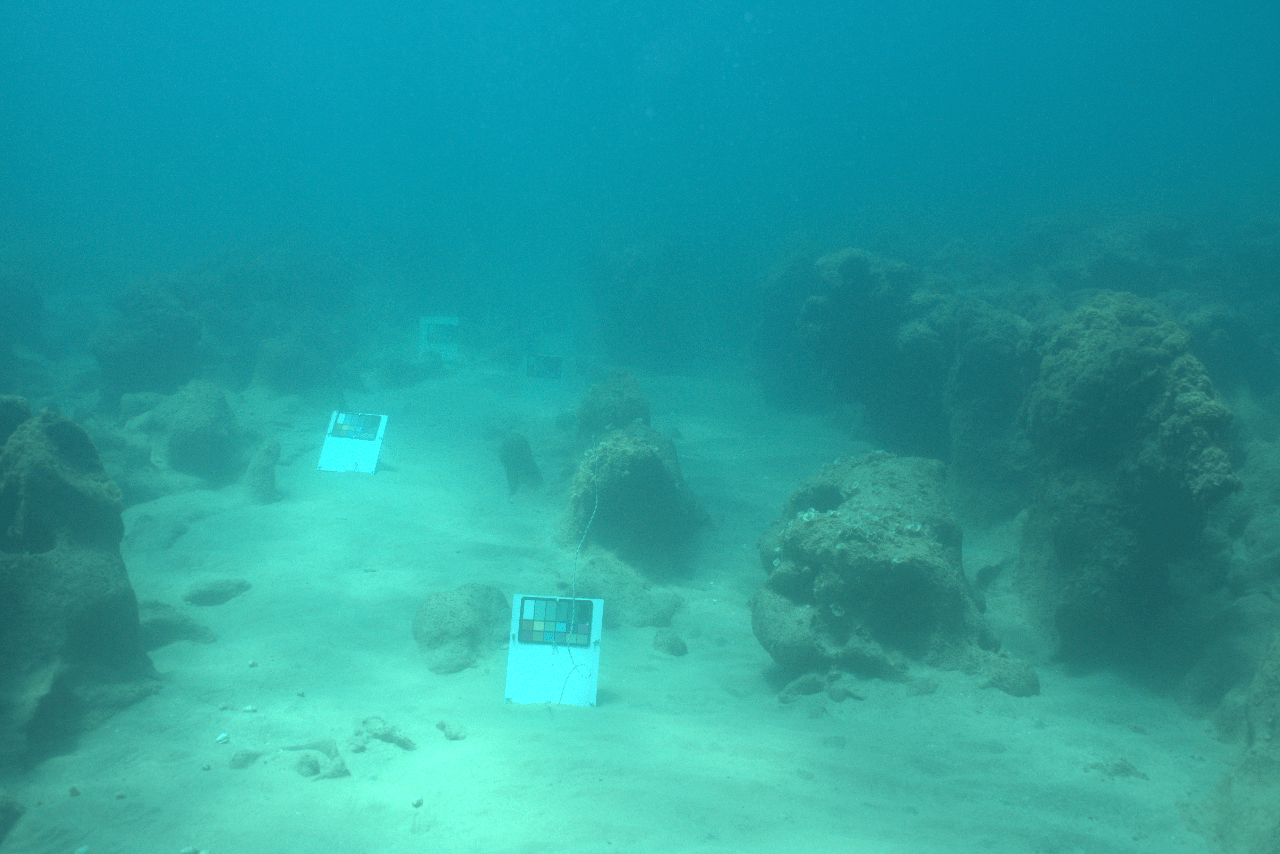} 
        \end{minipage}
        \captionsetup{labelformat=empty}
        \caption{Real underwater image}
        \vspace{-0.32cm}
    \end{subfigure}
    \hfill
    \hspace{-0.25cm}
    \begin{subfigure}[b]{0.120\textwidth}
        \centering
        \begin{minipage}[b][3cm][b]{1.0\textwidth}
          \vspace*{\fill}
          \centering
          \includegraphics[width=1.0\linewidth]{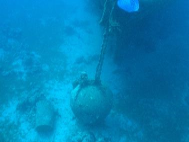} 
          \par\vspace{1mm}
          \includegraphics[width=1.0\linewidth]{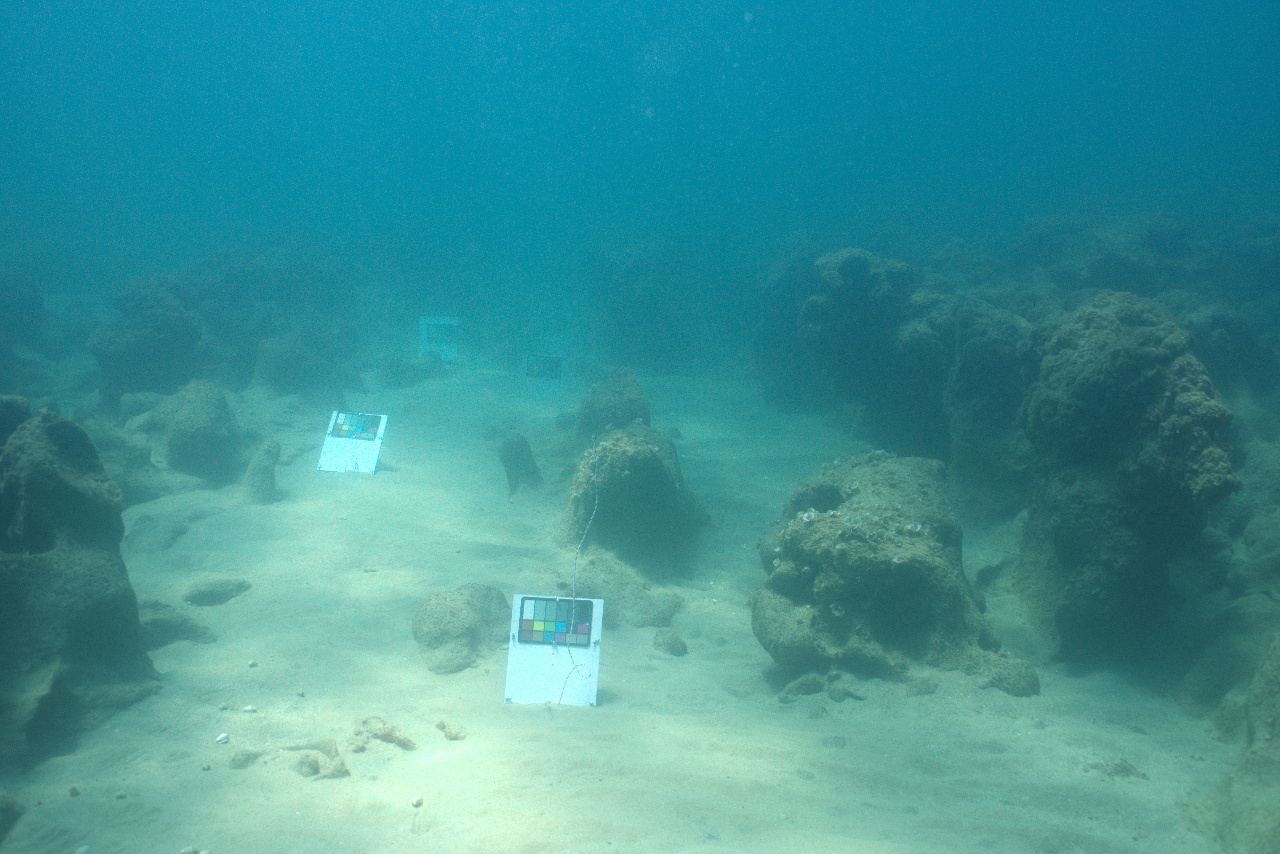} 
        \end{minipage}
        \captionsetup{labelformat=empty}
        \caption{UIBLA \cite{UIBLA}}
    \end{subfigure}
    \hfill
    \hspace{-0.25cm}
    \begin{subfigure}[b]{0.120\textwidth}
        \centering
        \begin{minipage}[b][3cm][b]{1.0\textwidth}
          \vspace*{\fill}
          \centering
          \includegraphics[width=1.0\linewidth]{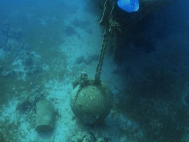} 
          \par\vspace{1mm}
          \includegraphics[width=1.0\linewidth]{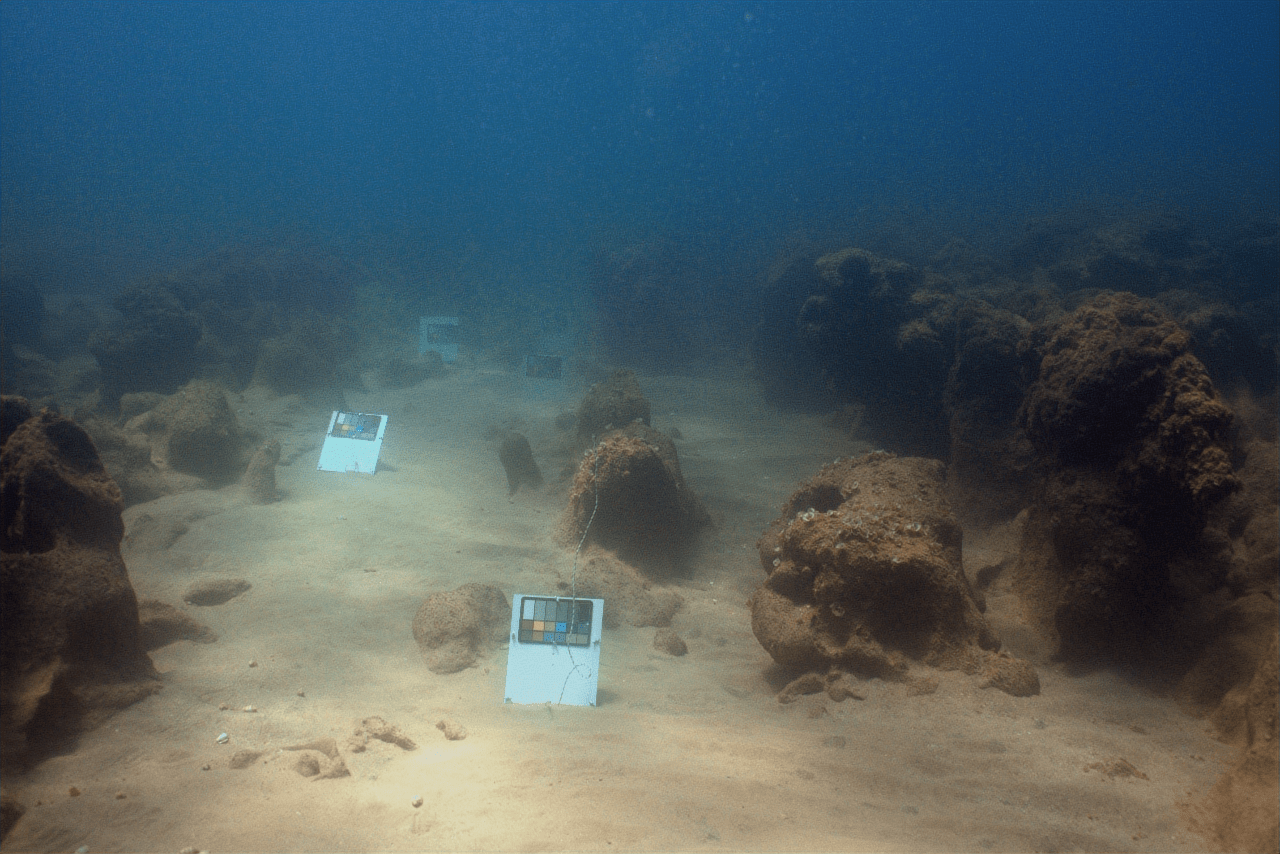} 
        \end{minipage}
        \captionsetup{labelformat=empty}
        \caption{Water-Net \cite{waternet}}
    \end{subfigure}
    \hfill
    \hspace{-0.25cm}
    \begin{subfigure}[b]{0.120\textwidth}
        \centering
        \begin{minipage}[b][3cm][b]{1.0\textwidth}
          \vspace*{\fill}
          \centering
          \includegraphics[width=1.0\linewidth]{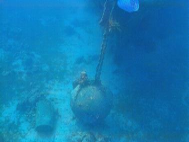} 
          \par\vspace{1mm}
          \includegraphics[width=1.0\linewidth]{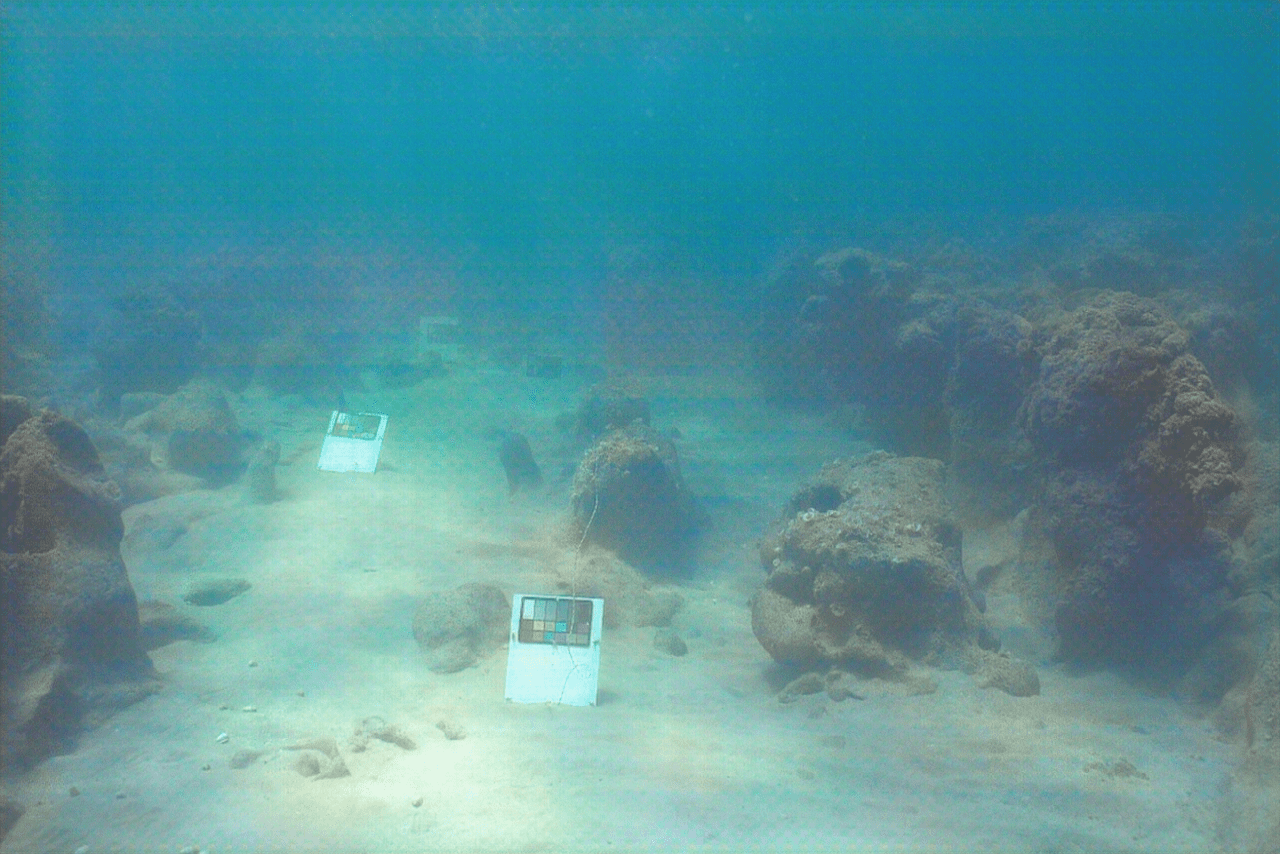} 
        \end{minipage}
        \captionsetup{labelformat=empty}
        \caption{FUnIE-GAN \cite{FUnIE-GAN}}
    \end{subfigure}
    \hfill
    \hspace{-0.25cm}
    \begin{subfigure}[b]{0.120\textwidth}
        \centering
        \begin{minipage}[b][3cm][b]{1.0\textwidth}
          \vspace*{\fill}
          \centering
          \includegraphics[width=1.0\linewidth]{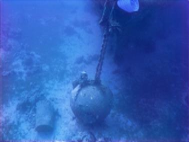} 
          \par\vspace{1mm}
          \includegraphics[width=1.0\linewidth]{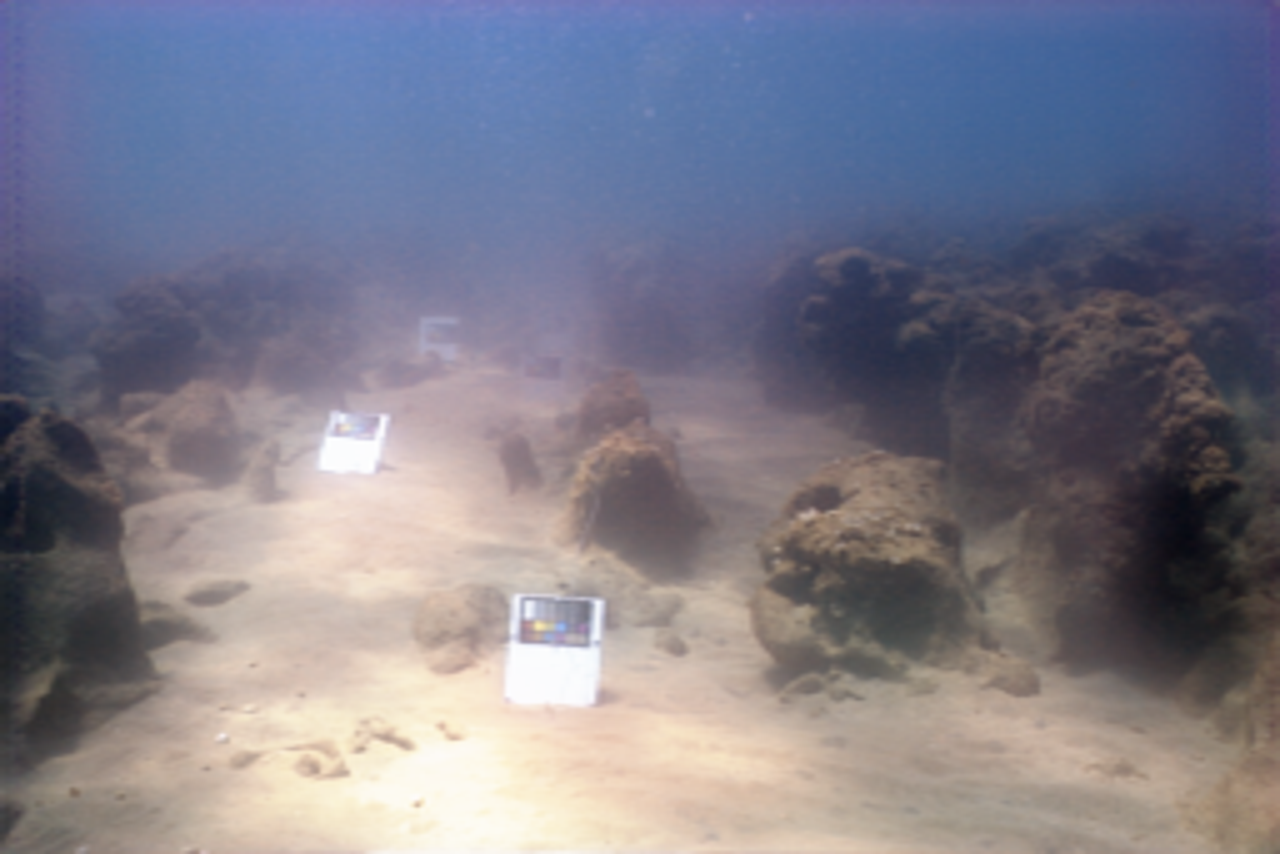} 
        \end{minipage}
        \captionsetup{labelformat=empty}
        \caption{UWGAN \cite{UWGAN}}
    \end{subfigure}
    \hfill
    \hspace{-0.25cm}
    \begin{subfigure}[b]{0.120\textwidth}
        \centering
        \begin{minipage}[b][3cm][b]{1.0\textwidth}
          \vspace*{\fill}
          \centering
          \includegraphics[width=1.0\linewidth]{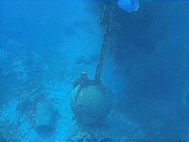} 
          \par\vspace{1mm}
          \includegraphics[width=1.0\linewidth]{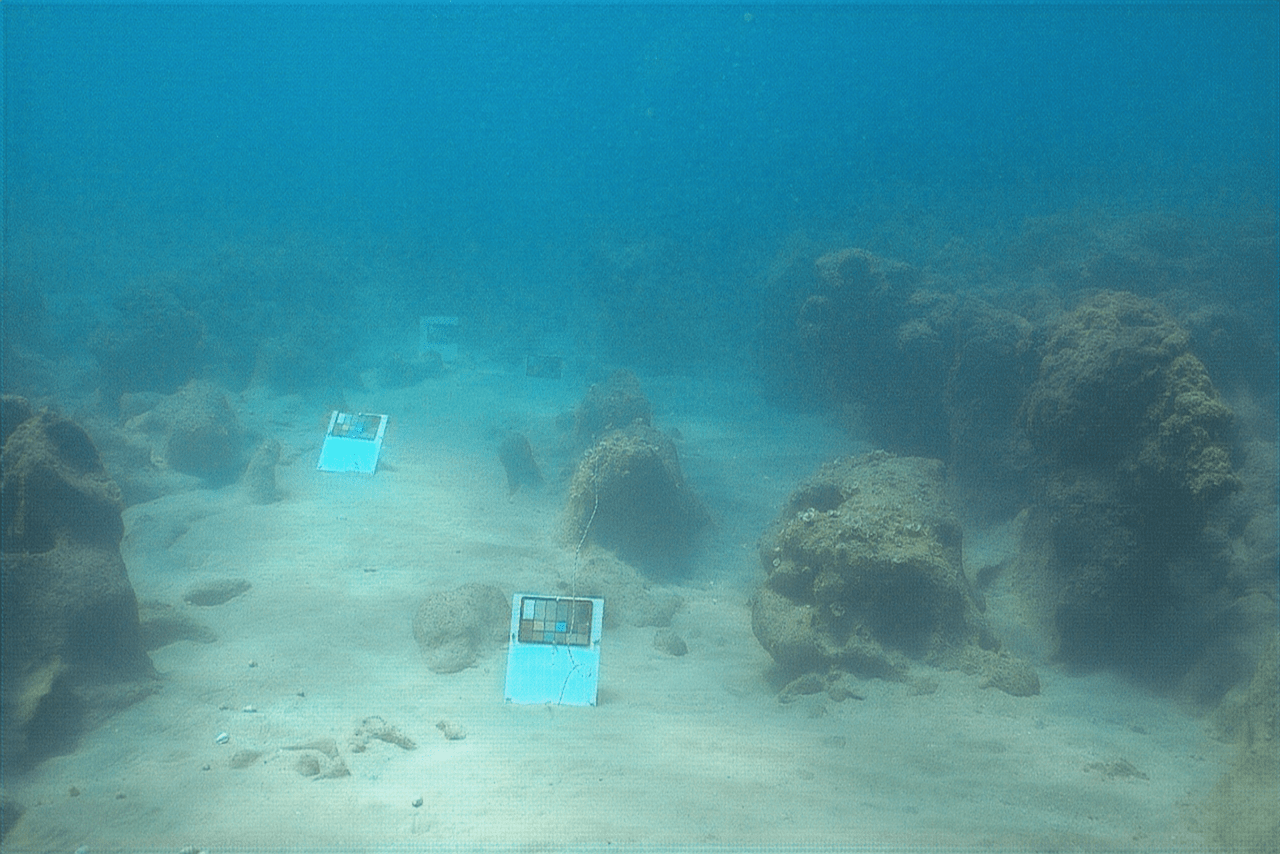} 
        \end{minipage}
        \captionsetup{labelformat=empty}
        \caption{UGAN \cite{UGAN}}
    \end{subfigure}
    \hfill
    \hspace{-0.25cm}
   \begin{subfigure}[b]{0.120\textwidth}
       \centering
        \begin{minipage}[b][3cm][b]{1.0\textwidth}
          \vspace*{\fill}
          \centering
          \includegraphics[width=1.0\linewidth]{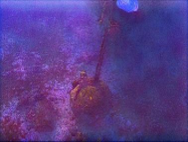} 
          \par\vspace{1mm}
          \includegraphics[width=1.0\linewidth]{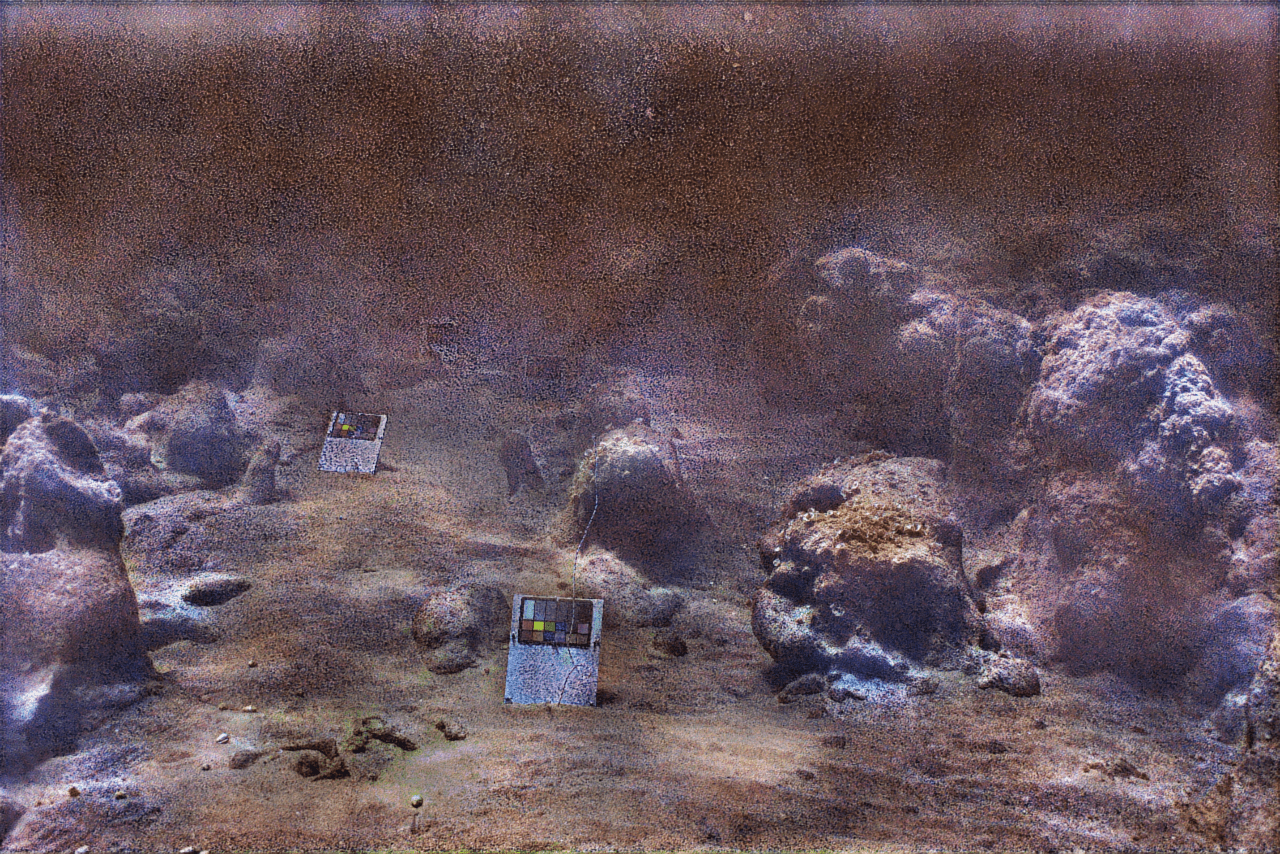} 
        \end{minipage} 
        \captionsetup{labelformat=empty}
        \caption{UIE-DAL \cite{UIE-DAL}}
    \end{subfigure}
    \hfill
    \hspace{-0.25cm}
    \begin{subfigure}[b]{0.120\textwidth}
        \centering
        \begin{minipage}[b][3cm][b]{1.0\textwidth}
          \vspace*{\fill}
          \centering
          \includegraphics[width=1.0\linewidth]{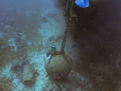} 
          \par\vspace{1mm}
          \includegraphics[width=1.0\linewidth]{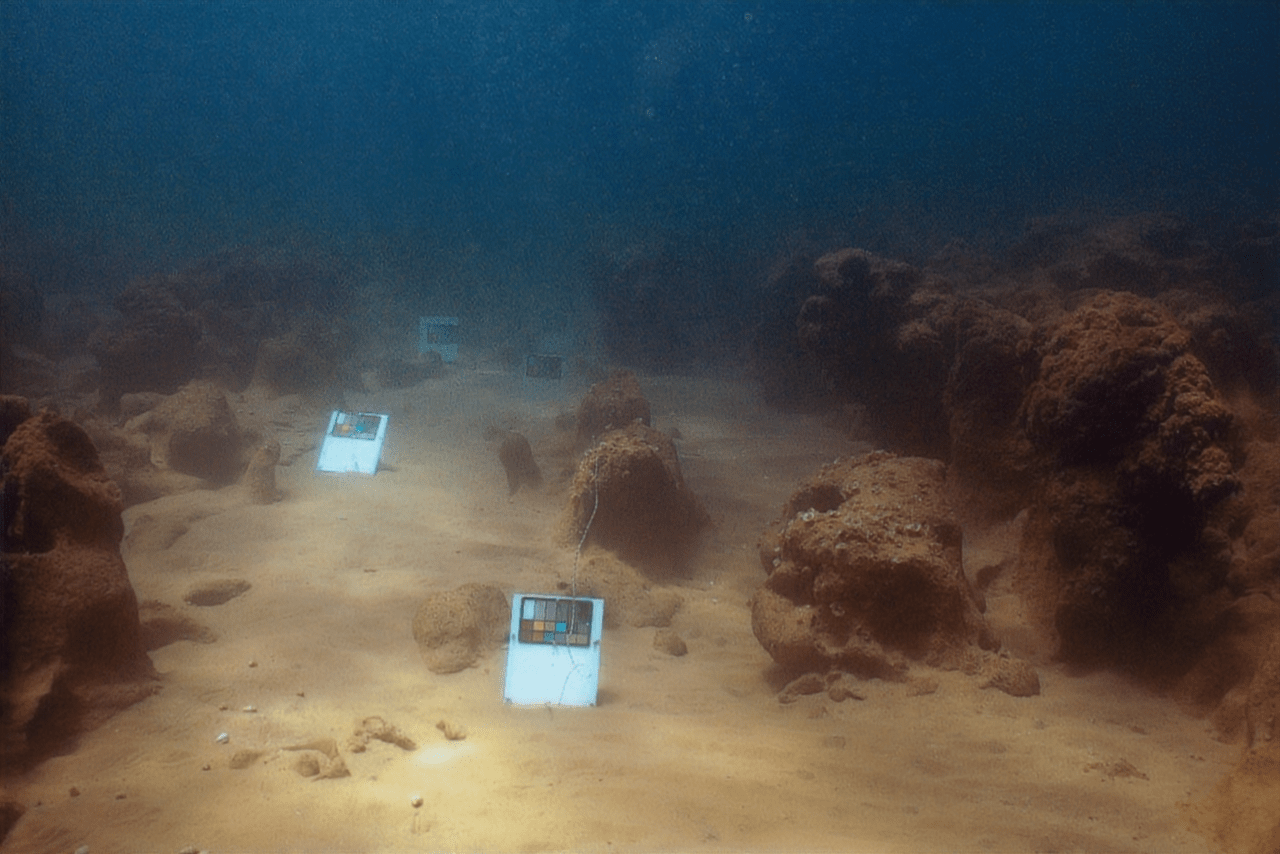} 
        \end{minipage}
        \captionsetup{labelformat=empty}
        \caption{UIESS (Ours)}
    \end{subfigure}
  \caption{Visual comparisons of real-world data from SQUID dataset.}
  \label{fig_real_quan_SQUID}
\end{figure*}

\subsection{Quantity Comparison}
\noindent Fig \ref{fig_syn_quan} depict the image enhancement result on synthetic data. The model based method, i.e. UIBLA, prone to under-enhanced over-enhanced while the degradation level is diverse. UIE-DAL generate high saturation result but loss fidelity to ground truth, noting that UIE-DAL is class of domain adaptation for underwater image enhancement method, thus it should perform well on no matter synthetic or real-world data. The others learning based fail to enhanced the medium level degradation image shown in top-row of Fig \ref{fig_syn_quan}, and remain purplish color-cast or under-enhanced, this is because these methods encode content and style into one latent, and make the model hard to distinguish the degradation level, i.e. style, to perform appropriate enhanced level. For the second-row of the image, we elaborate a slight degradation example, in this case, almost all of comparisons perform well. We emphasize the area of red bounding box and shown in third-row, our UIESS can generate more closer to ground truth result for dark area, while others method remain greenish distortion. For the rock area of third-row image, UIBLA over-enhanced; saturation of Water-Net, FUnIE-GAN and UGAN generated are lower than ground truth. The global appearance of generated results of UWGAN and UIE-DAL are different to ground truth obviously.       

For real-world data, visual comparison on UIEB and EUVP are shown in Fig \ref{fig_real_quan_UIEB} and Fig \ref{fig_real_quan_EUVP}. As mention in synthetic image result, each algorithms suffer from same deficiency, meanwhile, the domain gap problem make the others method more unstable and fail to enhance in many examples. In contrast, our model can remove the color-shift successfully and achieve clear and high saturation visual pleasing result for whether synthesis or real-world data and different degradation level. We further provide visual comparison on Sea-thru and SQUID datasets. SQUID consists many deep sea images, which contain heavy color cast, hazy effect and marine snow, on the other hand, most of image in Sea-thru dataset took in shallow sea, which the images do not have hazy effect. The result are shown in Fig \ref{fig_real_quan_Seathru} and Fig \ref{fig_real_quan_SQUID}. For Sea-thru dataset, almost all methods achieve reasonable result for top-row image, except UIBLA, which is under-enhanced, and Water-Net generate blueish color cast. Compare to another domain adaptation for underwater image enhance work, i.e. UIE-DAL, that aim to minimize the discrepancy of encoded latent from different domain, UIE-DAL still fail to generalize to real-world domain. In contrary, our model remove the underwater degradation successfully in both images, hence, we argue that using content style disentangling on domain adaptation for underwater image enhancement is more effective. The top-row image of SQUID dataset in Fig \ref{fig_real_quan_SQUID} demonstrate the severely color shift and hazy effect example, while others comparison method fail to enhance and output low contrast and more serious color cast degraded images, although our model cannot generate clear high quality result, the model reduce the blueish color cast and hazy effect significantly.

\subsection{Cross Domain Image-to-Image Translation}
\begin{figure}[tbp!]
  \centering
  \begin{subfigure}[b]{0.155\linewidth}
    \centering
    \includegraphics[width=\linewidth]{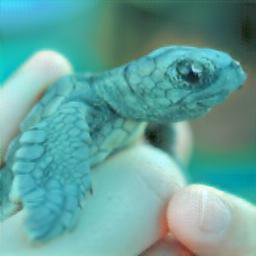}
    \caption{$I_{\scaleto{S}{3pt}}$}
  \end{subfigure}
  \begin{subfigure}[b]{0.155\linewidth}
    \centering
    \includegraphics[width=\linewidth]{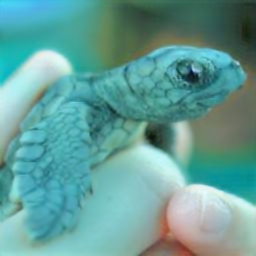}
    \caption{$I_{\scaleto{S \rightarrow S}{3pt}}$}
  \end{subfigure}
  \begin{subfigure}[b]{0.155\linewidth}
    \centering
    \includegraphics[width=\linewidth]{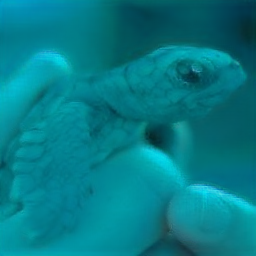}
    \caption{$I_{\scaleto{S \rightarrow R}{3pt}}$}
  \end{subfigure}
  \begin{subfigure}[b]{0.155\linewidth}
    \centering
    \includegraphics[width=\linewidth]{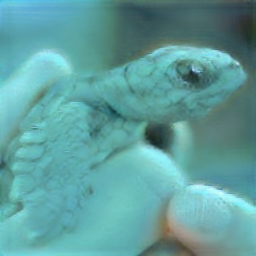}
    \caption{$I_{\scaleto{S \rightarrow R \rightarrow S}{3pt}}$}
  \end{subfigure}
  \begin{subfigure}[b]{0.155\linewidth}
    \centering
    \includegraphics[width=\linewidth]{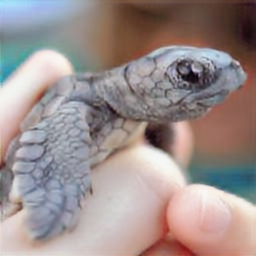}
    \caption{$I_{\scaleto{S \rightarrow C}{3pt}}$}
  \end{subfigure}
  \begin{subfigure}[b]{0.155\linewidth}
    \centering
    \includegraphics[width=\linewidth]{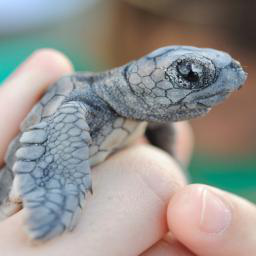}
    \caption{GT}
  \end{subfigure}
  
  \vspace{0.2cm}
  \begin{subfigure}[b]{0.180\linewidth}
    \centering
    \includegraphics[width=\linewidth]{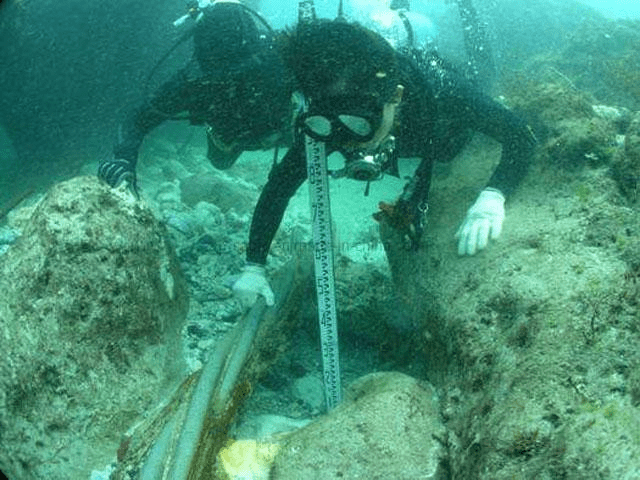}
    \caption{$I_{\scaleto{R}{3pt}}$}
  \end{subfigure}
  \begin{subfigure}[b]{0.180\linewidth}
    \centering
    \includegraphics[width=\linewidth]{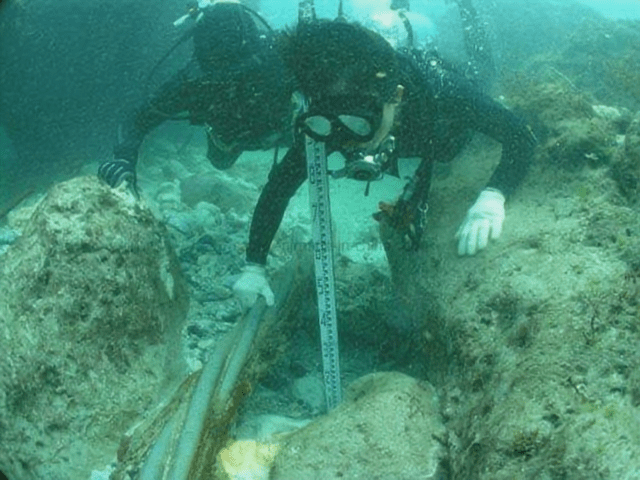}
    \caption{$I_{\scaleto{R \rightarrow R}{3pt}}$}
  \end{subfigure}
  \begin{subfigure}[b]{0.180\linewidth}
    \centering
    \includegraphics[width=\linewidth]{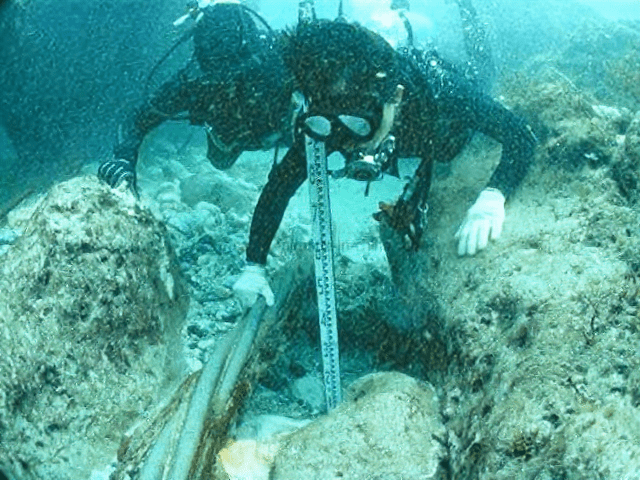}
    \caption{$I_{\scaleto{R \rightarrow S}{3pt}}$}
  \end{subfigure}
  \begin{subfigure}[b]{0.180\linewidth}
    \centering
    \includegraphics[width=\linewidth]{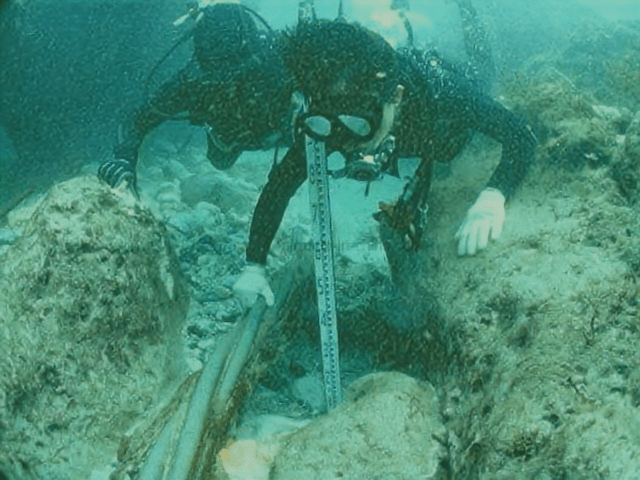}
    \caption{$I_{\scaleto{R \rightarrow S \rightarrow R}{3pt}}$}
  \end{subfigure}
  \begin{subfigure}[b]{0.180\linewidth}
    \centering
    \includegraphics[width=\linewidth]{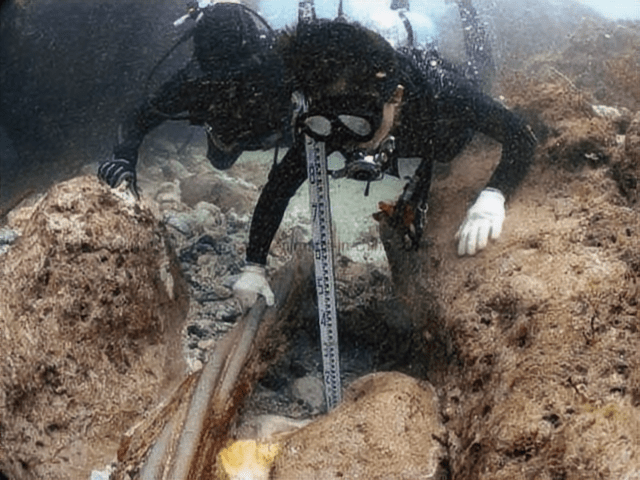}
    \caption{$I_{\scaleto{R \rightarrow C}{3pt}}$}
  \end{subfigure}
  \captionsetup{justification=raggedright,singlelinecheck=true}
  \caption{Our framework can perform cross domain image-to-image translation and enhancement simultaneously. (a) input synthesis underwater image. (b) reconstructed underwater image. (c) synthesis-to-real. (d) cycle-consistency image (e) enhanced image. (f) Ground truth. (g) - (h) for real-world example.}
\label{fig_translation}
\end{figure}

\noindent Our model can perform not only image enhancement, but also cross domain image-to-image translation, which act as real-world style underwater image synthesis. Noting that the cross domain image-to-image translation in image space is the additional benefit in the design, our goal is to explore representative latent space for synthesis, real-world underwater and clean latent, then perform latent space enhancement. To evaluate the representation of encoded style latent, we can input different style latent with same content into generator and visualize in image space, that is, to perform cross domain image-to-image translation. We demonstrate the domain translation result in Fig \ref{fig_translation} for synthesis and real image and take a real-world image as an example, given real-world underwater image $I_R$ in Fig \ref{fig_translation} (g), we can input different style latent to perform image reconstruction, domain translation and image enhancement and shown in Fig \ref{fig_translation} (h), (i) and (k). Fig \ref{fig_translation} (a)-(f) present a synthesis image example. We emphasize that under content style disentangling fashion, the latent space is more interpretable and operational, we will dive deeper to latent space analysis in next sub-section.

% translation

% A synthesis underwater image I2I and enhancement example, from left to right are input image, Syn$\rightarrow$Syn, Syn$\rightarrow$Real, Syn$\rightarrow$Real$\rightarrow$Syn, Syn$\rightarrow$Clean, GT.
%A real underwater image I2I and enhancement example, from left to right are input image, Real$\rightarrow$Real, Real$\rightarrow$Syn, Real$\rightarrow$Syn$\rightarrow$Real, Real$\rightarrow$Clean

%T-SNE
\begin{figure}[htbp!]
  \centering
  \begin{subfigure}[b]{0.23\textwidth}
    \centering
    \includegraphics[width=1.0\linewidth]{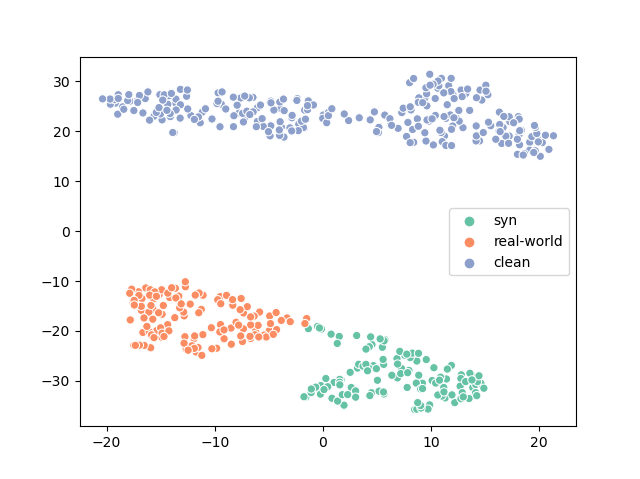}
    \caption{}
  \end{subfigure}
  \begin{subfigure}[b]{0.23\textwidth}
    \centering
    \includegraphics[width=1.0\linewidth]{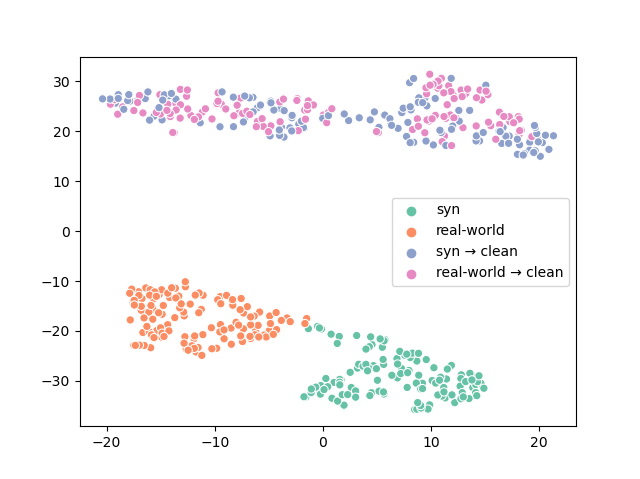}
    \caption{}
  \end{subfigure}
  \captionsetup{justification=raggedright,singlelinecheck=true}
  \caption{t-SNE visualization of style latent from different domains. (a) shown the synthesis, real-world underwater and clean latent in different color, clean latent include syn$\rightarrow$clean and real$\rightarrow$clean. To evaluate if these two different clean latents share same distribution, we split these two source for different color and visualize in (b)}
  \label{fig_tsne}
\end{figure}
%
%interpolate
\begin{figure*}[htbp!]
  \centering
  \begin{subfigure}[b]{1.0\textwidth}
    \centering
    \includegraphics[width=\textwidth]{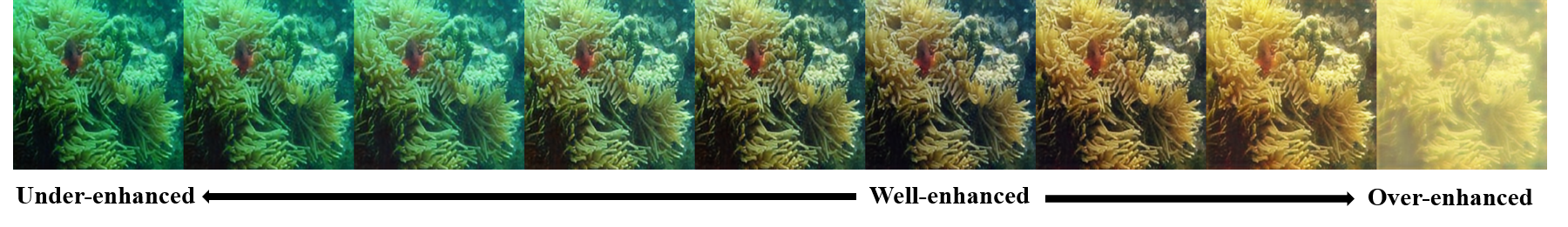}
    \caption{}
  \end{subfigure}
  \begin{subfigure}[b]{1.0\textwidth}
    \centering
    \includegraphics[width=\textwidth]{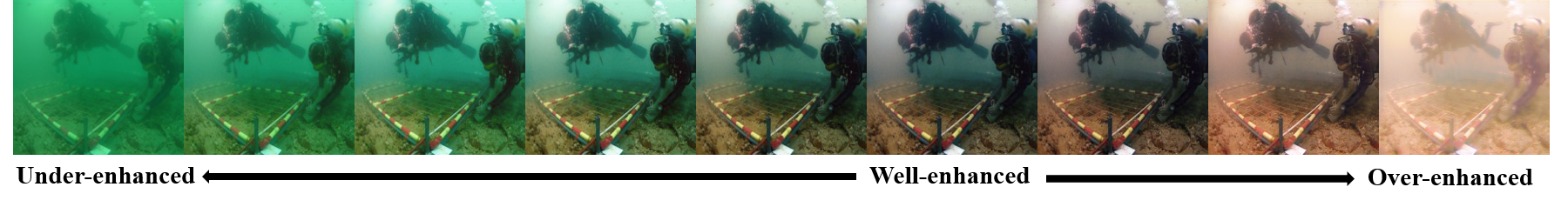}
    \caption{}
  \end{subfigure}
  \captionsetup{justification=raggedright,singlelinecheck=true}
  \caption{The examples of style latent manipulation, by given different $\alpha$ in Eq. \ref{eq_latent}, our model can achieve different enhancement level. (a) present a synthesis underwater example and (b) present a real-world underwater example}
  \label{fig_interpolate}
\end{figure*}

\subsection{Latent Analysis}
\noindent To learn more about latent space, we first evaluate the assumption of the synthesis, real-world underwater, and clean style latent belong to three domains. We visualize the style latent with t-SNE, the result are shown in Fig \ref{fig_tsne}, and as Fig \ref{fig_tsne} shown, the latent are properly clustered into three cluster as expected. Since the appearance of synthesis and real underwater image is more similar, the distance of these two cluster is closer than each one to clean cluster. We also examine that if the latent from synthesis underwater to clean $Z^{\scaleto{S}{4pt}}_{\scaleto{S \rightarrow C}{4pt}}$ and latent from real-world underwater to clean $Z^{\scaleto{S}{4pt}}_{\scaleto{R \rightarrow C}{4pt}}$ belongs to same distribution, we plot these two kind of clean latent in different color in Fig \ref{fig_tsne} (b). As the figure shown, latent of these two kind of clean latent mix together in same distribution.

Different from previous works \cite{UIE-DAL, physical, TUDA} of domain adaptation for underwater image enhancement method, which could not manipulate or interpret encoded latent, our model can perform latent manipulation thank to meaningful and operational latent space, and further obtain different level of enhanced result, which can act as a user interact parameter to adjust the level of enhancement. \citet{interfacegan} has discovered that for GAN-based model, a meaningful direction can be found for semantic editing by latent linear arithmetic, to figure out the target direction is a non-trivial topic \cite{interfacegan}. However, we can identify the direction of editing enhancement level simply for the vector formed by original style latent and enhanced one. Noting that the prior works could not perform same operation due to the encoded latent is mixed up with content and style. Enhancement level adjustment can be implement using following formula to manipulate style latent: 
\begin{equation}
    Z^{\scaleto{S}{4pt}}_{\varepsilon\rightarrow C} = Z^{\scaleto{S}{4pt}}_\varepsilon + \alpha \times (Z^{\scaleto{S}{4pt}}_{\varepsilon \rightarrow \scaleto{C}{4pt}} - Z^{\scaleto{S}{4pt}}_{\varepsilon}), \varepsilon=\{S, R\}
    \label{eq_latent}
\end{equation}
where $\alpha$ is weighting of enhanced level. For $\alpha \in (0,1)$, Eq. \ref{eq_latent} perform linear interpolation of original and enhanced style latent, and intermediate result from input image to well-enhanced image can be obtained. For $\alpha \textgreater 1$, it would generate over-enhanced result and $\alpha \textless 0$ would produce heavier degraded images. Fig \ref{fig_interpolate} demonstrate the latent manipulation of synthesis underwater image and real underwater image, the image continuous change between different enhanced level from input degraded image (i.e. under-enhanced image) to over-enhanced.

\subsection{Model Complexity Analysis}
\noindent To evaluate the model complexity of proposed method, we report the number of parameters and run time speed in Table \ref{table_model_complexity}, the run time speed is recorded for average run time of 515 256 $\times$ 256 images. Most of learning based method achieve real time processing, compare with another domain adaptation methods, i.e. UIE-DAL, proposed method use less parameters and has faster processing speed and achieve real-time processing with 148 FPS.

% model complexity
\begin{table}[htbp]
\captionsetup{justification=raggedright,singlelinecheck=true}
\caption{Model complexity evaluation. We list number of parameters and run time speed to compare the efficiency of different methods}
\begin{center}
\begin{tabular}{|c|c|c|}
\hline
\textbf{Methods} & \textbf{\textit{Para. (M)}} $\downarrow$ & \textbf{\textit{Time (ms)}} $\downarrow$\\
\hline
UIBLA& - & 5300  \\
Water-Net & 1.09 & 63.135 \\
FUnIE-GAN & 7.02 & 4.295 \\
UWGAN & 1.93 &  5.359\\
UGAN & 54.40 & 6.127 \\
UIE-DAL & 13.40 &  11.197 \\
Ours & 4.26 & 6.795 \\
\hline
\end{tabular}
\label{table_model_complexity}`
\end{center}
\end{table}

\subsection{Ablation Study}
\noindent We conduct an ablation study on loss function by removing some term in total loss function and show the effectiveness of removed term. As the result shown in Table \ref{table_ablation}, by removing cycle-consistency loss, L1 loss, SSIM loss and perceptual loss, the performance drop. Especially by removing SSIM loss, the performance drop dramatically because SSIM loss provide the structural and contrast information while model training, which is more crucial then pixel fidelity to human visual system and colorfulness evaluation. The experiment is conducted on UIEB dataset.

\begin{table}[htbp]
\captionsetup{justification=raggedright,singlelinecheck=true}
\caption{Ablation study on our proposed method, we evaluate the effectiveness of loss function.}
\begin{center}
\begin{tabular}{|c|c|c|}
\hline
\textbf{Methods} & \textbf{\textit{UIQM}} $\uparrow$ & \textbf{\textit{UCIQE}} $\uparrow$\\
\hline
w/o Cycle-consistency loss & 3.8103 & 0.5853  \\
w/o L1 loss & 3.8523 & 0.5801 \\
w/o SSIM loss & 3.7522 & 0.5763 \\
w/o perceptual loss & 3.7585 & 0.5880 \\
Ours & \textbf{3.9165} & \textbf{0.5950} \\
\hline
\end{tabular}
\label{table_ablation}`
\end{center}
\end{table}

\subsection{Generalization Study on Different Applications}
\begin{figure}[tbp!]
  \centering
  \begin{subfigure}[b]{0.240\linewidth}
    \centering
    \includegraphics[width=\linewidth]{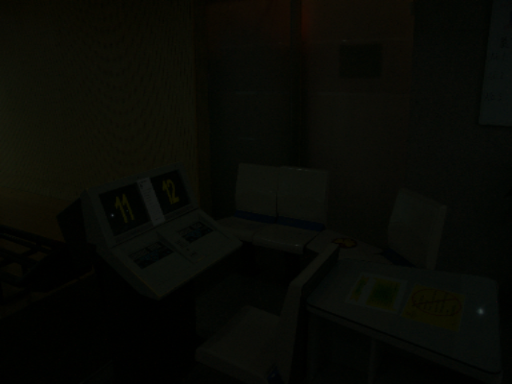}
    \captionsetup{labelformat=empty}
    \caption{Real low-light image}
  \end{subfigure}
  \begin{subfigure}[b]{0.240\linewidth}
    \centering
    \includegraphics[width=\linewidth]{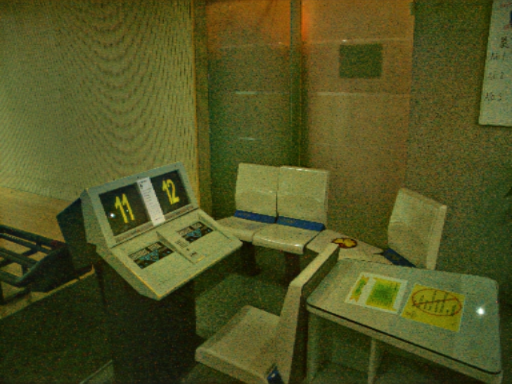}
    \captionsetup{labelformat=empty}
    \caption{EnlightenGAN \cite{enlightengan}}
  \end{subfigure}
  \begin{subfigure}[b]{0.240\linewidth}
    \centering
    \includegraphics[width=\linewidth]{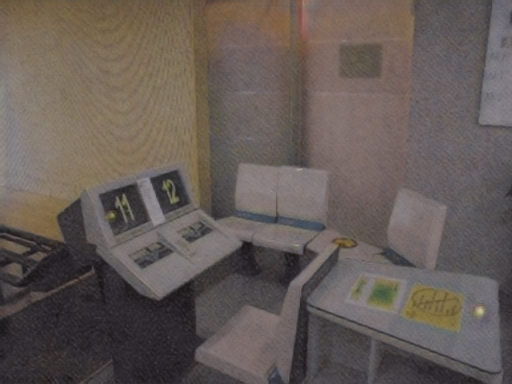}
    \captionsetup{labelformat=empty}
    \caption{Ours}
    \vspace{0.32cm}
  \end{subfigure}
  \begin{subfigure}[b]{0.240\linewidth}
    \centering
    \includegraphics[width=\linewidth]{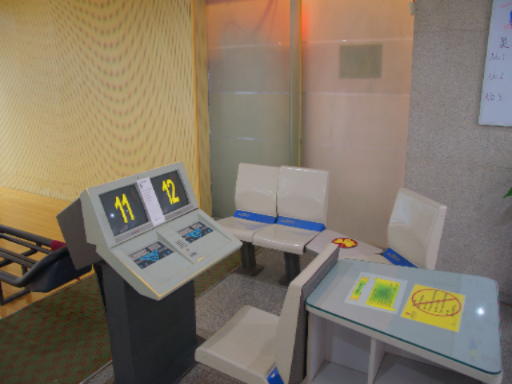}
    \captionsetup{labelformat=empty}
    \caption{GT}
    \vspace{0.32cm}
  \end{subfigure}
  
  \begin{subfigure}[b]{0.240\linewidth}
    \centering
    \includegraphics[width=\linewidth]{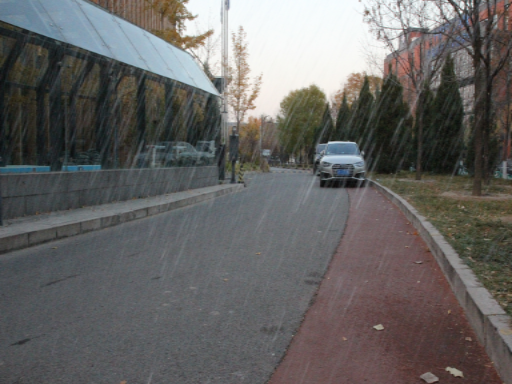}.
    \captionsetup{labelformat=empty}
    \caption{Real rain image}
  \end{subfigure}
  \begin{subfigure}[b]{0.240\linewidth}
    \centering
    \includegraphics[width=\linewidth]{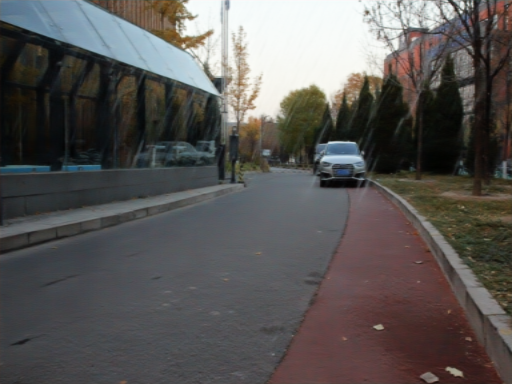}
    \captionsetup{labelformat=empty}
    \caption{DID-MDN \cite{did_mdn}}
  \end{subfigure}
  \begin{subfigure}[b]{0.240\linewidth}
    \centering
    \includegraphics[width=\linewidth]{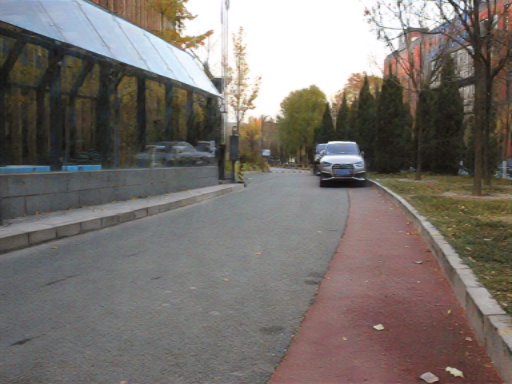}
    \captionsetup{labelformat=empty}
    \caption{Ours}
    %\vspace{0.32cm}
  \end{subfigure}
  \begin{subfigure}[b]{0.240\linewidth}
    \centering
    \includegraphics[width=\linewidth]{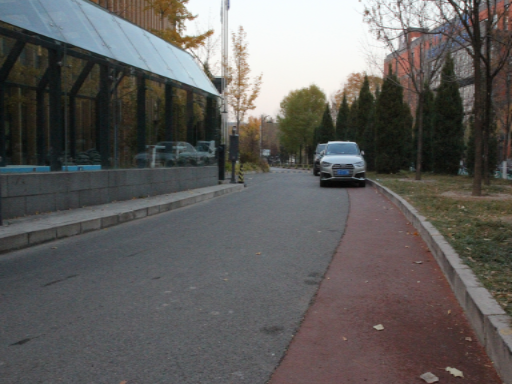}
    \captionsetup{labelformat=empty}
    \caption{GT}
    %\vspace{0.32cm}
  \end{subfigure}
  
  \begin{subfigure}[b]{0.240\linewidth}
    \raggedleft
    \includegraphics[width=\linewidth]{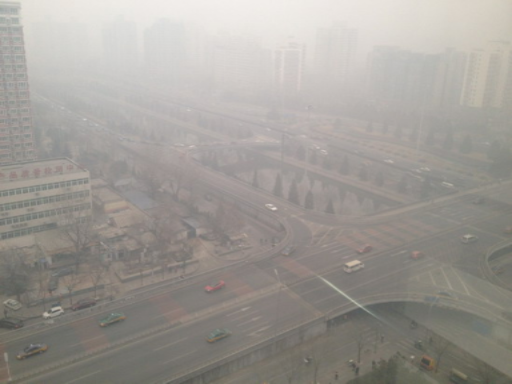}
    \captionsetup{labelformat=empty}
    \caption{Real hazy image}
  \end{subfigure}
  \begin{subfigure}[b]{0.240\linewidth}
    \includegraphics[width=\linewidth]{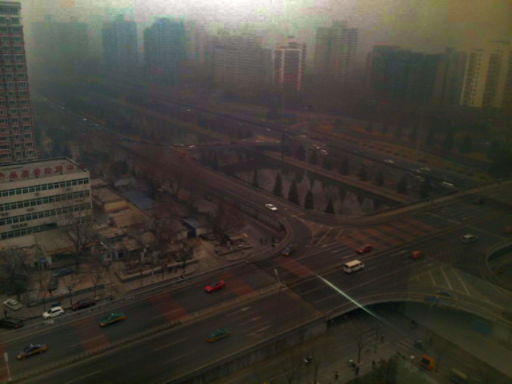}
    \captionsetup{labelformat=empty}
    \caption{DCP \cite{DCP}}
    %\vspace{0.32cm}
  \end{subfigure}
  \begin{subfigure}[b]{0.240\linewidth}
    \includegraphics[width=\linewidth]{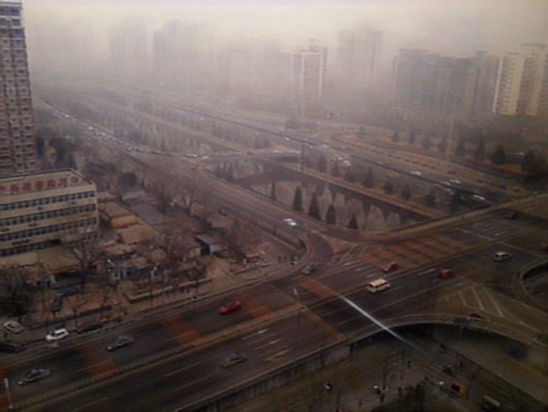}
    \captionsetup{labelformat=empty}
    \caption{Ours}
    %\vspace{0.32cm}
  \end{subfigure}
  \begin{subfigure}[b]{0.240\linewidth}
    \includegraphics[width=\linewidth]{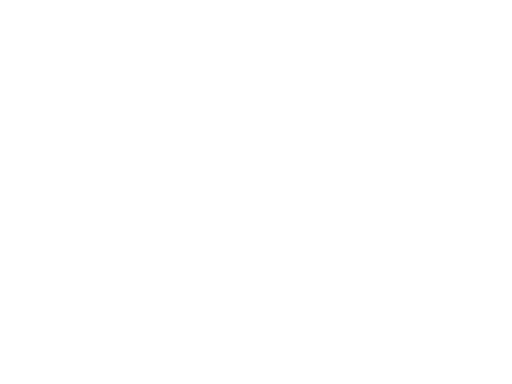}
    \captionsetup{labelformat=empty}
    \caption{}
    %\vspace{0.32cm}
  \end{subfigure}
  
  \captionsetup{justification=raggedright,singlelinecheck=true}
  \caption{Generalization study on different low-level vision task, include low-light enhancement, image derain streak and image dehazing, our model achieve competitive result.}
\label{fig_generalization}
\end{figure}

\noindent Our model is designed under a general perspective, and has potential to generalize to different low-level image enhancement task. We conduct several low-level vision task on proposed method to examine the generalization ability, include low-light enhancement, image derain streak and image dehazing. We retrain the model with real and synthetic dataset for corresponding task. For low-light enhancement, we have visual comparison with EnlightenGAN \cite{enlightengan}, and for derain streak and dehazing, we compare with DID-MDN \cite{did_mdn} and DCP \cite{DCP}. As the Fig \ref{fig_generalization} shown, our model can generalize to different low-level vision tasks, especially for global degradation removal, and achieve competitive performance even compare with methods proposed in recent years. We suppose the proposed method might be suitable to the benchmark for future domain adaptation for image enhancement research.  

% generalization study

\section{Conclusion}  
\noindent In this paper, we proposed a novel domain adaptation framework for underwater image enhancement via content and separation, our model can perform image-to-image translation and enhancement simultaneously and the model can be trained in end-to-end manner. Different to previous works, our model also can perform latent manipulation to modulate different enhancement level. Experiment on various synthesis and real-world datasets demonstrate our proposed method can achieve favorably against the state-of-the-art algorithms. The proposed framework can also generalize to different low-level vision tasks and has the potential serve as benchmark for future domain adaptation for image enhancement research. Possible future work include enhance the multiple degradation of underwater image in different semantic manner, under this setting, latent manipulation can be performed for each degradation and activate more flexible user interact interface. 

\bibliographystyle{unsrtnat}
%\nocite{*}
{\footnotesize
\bibliography{ref}

\begin{thebibliography}{44}
\providecommand{\natexlab}[1]{#1}
\providecommand{\url}[1]{\texttt{#1}}
\expandafter\ifx\csname urlstyle\endcsname\relax
  \providecommand{\doi}[1]{doi: #1}\else
  \providecommand{\doi}{doi: \begingroup \urlstyle{rm}\Url}\fi

\bibitem[Drews et~al.(2016)Drews, Nascimento, Botelho, and
  Montenegro~Campos]{UDCP}
Paulo~L.J. Drews, Erickson~R. Nascimento, Silvia~S.C. Botelho, and
  Mario~Fernando Montenegro~Campos.
\newblock Underwater depth estimation and image restoration based on single
  images.
\newblock \emph{IEEE Computer Graphics and Applications}, 36\penalty0
  (2):\penalty0 24--35, 2016.

\bibitem[Peng et~al.(2018)Peng, Cao, and Cosman]{GDCP}
Yan-Tsung Peng, Keming Cao, and Pamela~C. Cosman.
\newblock Generalization of the dark channel prior for single image
  restoration.
\newblock \emph{IEEE Transactions on Image Processing}, 27\penalty0
  (6):\penalty0 2856--2868, 2018.

\bibitem[Li et~al.(2017)Li, Skinner, Eustice, and Johnson-Roberson]{waterGAN}
Jie Li, Katherine~A. Skinner, Ryan~M. Eustice, and Matthew Johnson-Roberson.
\newblock Watergan: Unsupervised generative network to enable real-time color
  correction of monocular underwater images.
\newblock \emph{IEEE Robotics and Automation Letters}, page 1–1, 2017.

\bibitem[Dudhane et~al.(2020)Dudhane, Hambarde, Patil, and
  Murala]{deepRestoration_SingalLetter2019}
Akshay Dudhane, Praful Hambarde, Prashant Patil, and Subrahmanyam Murala.
\newblock Deep underwater image restoration and beyond.
\newblock \emph{IEEE Signal Processing Letters}, 27:\penalty0 675--679, 2020.

\bibitem[Fu and Cao(2020)]{underwaterGlobalLocal_Signalprocessing}
Xueyang Fu and Xiangyong Cao.
\newblock Underwater image enhancement with global-local networks and
  compressed-histogram equalization.
\newblock \emph{Signal Processing: Image Communication}, 86:\penalty0 115892,
  2020.

\bibitem[Wang et~al.(2021{\natexlab{a}})Wang, Zhou, Han, Zhu, and Yao]{UWGAN}
Nan Wang, Yabin Zhou, Fenglei Han, Haitao Zhu, and Jingzheng Yao.
\newblock Uwgan: Underwater gan for real-world underwater color restoration and
  dehazing, 2021{\natexlab{a}}.

\bibitem[Fabbri et~al.(2018)Fabbri, Islam, and Sattar]{UGAN}
Cameron Fabbri, Md~Jahidul Islam, and Junaed Sattar.
\newblock Enhancing underwater imagery using generative adversarial networks.
\newblock In \emph{2018 IEEE International Conference on Robotics and
  Automation (ICRA)}, pages 7159--7165, 2018.

\bibitem[Islam et~al.(2020)Islam, Xia, and Sattar]{FUnIE-GAN}
Md~Jahidul Islam, Youya Xia, and Junaed Sattar.
\newblock Fast underwater image enhancement for improved visual perception.
\newblock \emph{IEEE Robotics and Automation Letters}, 5\penalty0 (2):\penalty0
  3227--3234, 2020.

\bibitem[Li et~al.(2020{\natexlab{a}})Li, Guo, Ren, Cong, Hou, Kwong, and
  Tao]{waternet}
Chongyi Li, Chunle Guo, Wenqi Ren, Runmin Cong, Junhui Hou, Sam Kwong, and
  Dacheng Tao.
\newblock An underwater image enhancement benchmark dataset and beyond.
\newblock \emph{IEEE Transactions on Image Processing}, 29:\penalty0
  4376--4389, 2020{\natexlab{a}}.

\bibitem[Li et~al.(2020{\natexlab{b}})Li, Anwar, and Porikli]{NYU_syn}
Chongyi Li, Saeed Anwar, and Fatih Porikli.
\newblock Underwater scene prior inspired deep underwater image and video
  enhancement.
\newblock \emph{Pattern Recognition}, 98:\penalty0 107038, 2020{\natexlab{b}}.

\bibitem[Uplavikar et~al.(2019)Uplavikar, Wu, and Wang]{UIE-DAL}
Pritish~M Uplavikar, Zhenyu Wu, and Zhangyang Wang.
\newblock All-in-one underwater image enhancement using domain-adversarial
  learning.
\newblock In \emph{CVPR Workshops}, pages 1--8, 2019.

\bibitem[Zhou et~al.(2021)Zhou, Yan, and Li]{physical}
Yuan Zhou, Kangming Yan, and Xiaofeng Li.
\newblock Underwater image enhancement via physical-feedback adversarial
  transfer learning.
\newblock \emph{IEEE Journal of Oceanic Engineering}, pages 1--11, 2021.

\bibitem[Wang et~al.(2021{\natexlab{b}})Wang, Shen, Yu, Wang, Lin, and
  Xu]{TUDA}
Zhengyong Wang, Liquan Shen, Mei Yu, Kun Wang, Yufei Lin, and Mai Xu.
\newblock Domain adaptation for underwater image enhancement,
  2021{\natexlab{b}}.

\bibitem[Shao et~al.(2020)Shao, Li, Ren, Gao, and Sang]{DAdehazing}
Yuanjie Shao, Lerenhan Li, Wenqi Ren, Changxin Gao, and Nong Sang.
\newblock Domain adaptation for image dehazing.
\newblock In \emph{Proceedings of the IEEE/CVF Conference on Computer Vision
  and Pattern Recognition}, pages 2808--2817, 2020.

\bibitem[Jiang et~al.(2022)Jiang, Zhang, Bao, Zhao, Zhang, and Liu]{two_step}
Qun Jiang, Yunfeng Zhang, Fangxun Bao, Xiuyang Zhao, Caiming Zhang, and Peide
  Liu.
\newblock Two-step domain adaptation for underwater image enhancement.
\newblock \emph{Pattern Recognition}, 122:\penalty0 108324, 2022.

\bibitem[Ancuti et~al.(2012)Ancuti, Ancuti, Haber, and Bekaert]{fusion}
Cosmin Ancuti, Codruta~Orniana Ancuti, Tom Haber, and Philippe Bekaert.
\newblock Enhancing underwater images and videos by fusion.
\newblock In \emph{2012 IEEE Conference on Computer Vision and Pattern
  Recognition}, pages 81--88, 2012.

\bibitem[Iqbal et~al.(2010)Iqbal, Odetayo, James, Salam, and Talib]{GLC21}
Kashif Iqbal, Michael Odetayo, Anne James, Rosalina~Abdul Salam, and Abdullah
  Zawawi~Hj Talib.
\newblock Enhancing the low quality images using unsupervised colour correction
  method.
\newblock In \emph{2010 IEEE International Conference on Systems, Man and
  Cybernetics}, pages 1703--1709, 2010.

\bibitem[{Abdul Ghani} and {Mat Isa}(2015)]{GLC22}
Ahmad~Shahrizan {Abdul Ghani} and Nor~Ashidi {Mat Isa}.
\newblock Underwater image quality enhancement through integrated color model
  with rayleigh distribution.
\newblock \emph{Applied Soft Computing}, 27:\penalty0 219--230, 2015.

\bibitem[Fu et~al.(2014)Fu, Zhuang, Huang, Liao, Zhang, and Ding]{waternet33}
Xueyang Fu, Peixian Zhuang, Yue Huang, Yinghao Liao, Xiao-Ping Zhang, and
  Xinghao Ding.
\newblock A retinex-based enhancing approach for single underwater image.
\newblock In \emph{2014 IEEE International Conference on Image Processing
  (ICIP)}, pages 4572--4576, 2014.

\bibitem[McGlamery(1980)]{GLC25}
B.~L. McGlamery.
\newblock {A Computer Model For Underwater Camera Systems}.
\newblock In Seibert~Quimby Duntley, editor, \emph{Ocean Optics VI}, volume
  0208, pages 221 -- 231. International Society for Optics and Photonics, SPIE,
  1980.

\bibitem[Jaffe(1990)]{GLC26}
J.S. Jaffe.
\newblock Computer modeling and the design of optimal underwater imaging
  systems.
\newblock \emph{IEEE Journal of Oceanic Engineering}, 15\penalty0 (2):\penalty0
  101--111, 1990.

\bibitem[Akkaynak and Treibitz(2019)]{seathru}
Derya Akkaynak and Tali Treibitz.
\newblock Sea-thru: A method for removing water from underwater images.
\newblock In \emph{2019 IEEE/CVF Conference on Computer Vision and Pattern
  Recognition (CVPR)}, pages 1682--1691, 2019.

\bibitem[Zhang et~al.(2021{\natexlab{a}})Zhang, Sun, Wu, and Gu]{dugan}
Huiqing Zhang, Luyu Sun, Lifang Wu, and Ke~Gu.
\newblock Dugan: An effective framework for underwater image enhancement.
\newblock \emph{IET Image Processing}, 15\penalty0 (9):\penalty0 2010--2019,
  2021{\natexlab{a}}.

\bibitem[Liu et~al.(2020)Liu, Song, and Ding]{detection0}
Hong Liu, Pinhao Song, and Runwei Ding.
\newblock Towards domain generalization in underwater object detection.
\newblock In \emph{2020 IEEE International Conference on Image Processing
  (ICIP)}, pages 1971--1975, 2020.

\bibitem[Zhang et~al.(2021{\natexlab{b}})Zhang, Fang, and Ma]{detection2}
Boying Zhang, Jingzhe Fang, and Zi'ao Ma.
\newblock Underwater target recognition method based on domain adaptation.
\newblock In \emph{2021 4th International Conference on Pattern Recognition and
  Artificial Intelligence (PRAI)}, pages 452--455, 2021{\natexlab{b}}.

\bibitem[Zhang et~al.(2018)Zhang, Tang, Li, Guo, Zhang, Li, and Yan]{S3GAN}
Rui Zhang, Sheng Tang, Yu~Li, Junbo Guo, Yongdong Zhang, Jintao Li, and
  Shuicheng Yan.
\newblock Style separation and synthesis via generative adversarial networks.
\newblock New York, NY, USA, 2018. Association for Computing Machinery.
\newblock ISBN 9781450356657.

\bibitem[Huang et~al.(2018)Huang, Liu, Belongie, and Kautz]{MUNIT}
Xun Huang, Ming-Yu Liu, Serge Belongie, and Jan Kautz.
\newblock Multimodal unsupervised image-to-image translation.
\newblock In \emph{Proceedings of the European Conference on Computer Vision
  (ECCV)}, 2018.

\bibitem[Tulyakov et~al.(2018)Tulyakov, Liu, Yang, and Kautz]{MoCoGAN}
Sergey Tulyakov, Ming-Yu Liu, Xiaodong Yang, and Jan Kautz.
\newblock Mocogan: Decomposing motion and content for video generation.
\newblock In \emph{Proceedings of the IEEE Conference on Computer Vision and
  Pattern Recognition (CVPR)}, 2018.

\bibitem[Bousmalis et~al.(2016)Bousmalis, Trigeorgis, Silberman, Krishnan, and
  Erhan]{DSN}
Konstantinos Bousmalis, George Trigeorgis, Nathan Silberman, Dilip Krishnan,
  and Dumitru Erhan.
\newblock Domain separation networks.
\newblock \emph{Advances in neural information processing systems},
  29:\penalty0 343--351, 2016.

\bibitem[Ulyanov et~al.(2017)Ulyanov, Vedaldi, and Lempitsky]{IN}
Dmitry Ulyanov, Andrea Vedaldi, and Victor Lempitsky.
\newblock Improved texture networks: Maximizing quality and diversity in
  feed-forward stylization and texture synthesis.
\newblock In \emph{2017 IEEE Conference on Computer Vision and Pattern
  Recognition (CVPR)}, pages 4105--4113, 2017.

\bibitem[Huang and Belongie(2017)]{AdaIN}
Xun Huang and Serge Belongie.
\newblock Arbitrary style transfer in real-time with adaptive instance
  normalization.
\newblock In \emph{2017 IEEE International Conference on Computer Vision
  (ICCV)}, pages 1510--1519, 2017.

\bibitem[Park et~al.(2019)Park, Liu, Wang, and Zhu]{spade}
Taesung Park, Ming-Yu Liu, Ting-Chun Wang, and Jun-Yan Zhu.
\newblock Semantic image synthesis with spatially-adaptive normalization.
\newblock In \emph{Proceedings of the IEEE Conference on Computer Vision and
  Pattern Recognition}, 2019.

\bibitem[Wang et~al.(2018)Wang, Liu, Zhu, Tao, Kautz, and
  Catanzaro]{discriminator}
Ting-Chun Wang, Ming-Yu Liu, Jun-Yan Zhu, Andrew Tao, Jan Kautz, and Bryan
  Catanzaro.
\newblock High-resolution image synthesis and semantic manipulation with
  conditional gans.
\newblock In \emph{Proceedings of the IEEE Conference on Computer Vision and
  Pattern Recognition (CVPR)}, 2018.

\bibitem[Mao et~al.(2017)Mao, Li, Xie, Lau, Wang, and Paul~Smolley]{LSGAN}
Xudong Mao, Qing Li, Haoran Xie, Raymond~Y.K. Lau, Zhen Wang, and Stephen
  Paul~Smolley.
\newblock Least squares generative adversarial networks.
\newblock In \emph{Proceedings of the IEEE International Conference on Computer
  Vision (ICCV)}, 2017.

\bibitem[Seif and Androutsos(2018)]{loss_ssim_L1_0}
George Seif and Dimitrios Androutsos.
\newblock Edge-based loss function for single image super-resolution.
\newblock In \emph{2018 IEEE International Conference on Acoustics, Speech and
  Signal Processing (ICASSP)}, pages 1468--1472, 2018.

\bibitem[Zhao et~al.(2017)Zhao, Gallo, Frosio, and Kautz]{loss_ssim_L1_1}
Hang Zhao, Orazio Gallo, Iuri Frosio, and Jan Kautz.
\newblock Loss functions for image restoration with neural networks.
\newblock \emph{IEEE Transactions on Computational Imaging}, 3\penalty0
  (1):\penalty0 47--57, 2017.

\bibitem[Peng and Cosman(2017)]{UIBLA}
Yan-Tsung Peng and Pamela~C. Cosman.
\newblock Underwater image restoration based on image blurriness and light
  absorption.
\newblock \emph{IEEE Transactions on Image Processing}, 26\penalty0
  (4):\penalty0 1579--1594, 2017.

\bibitem[Wang et~al.(2019)Wang, Song, Fortino, Qi, Zhang, and
  Liotta]{UIBLA_code}
Yan Wang, Wei Song, Giancarlo Fortino, Li-Zhe Qi, Wenqiang Zhang, and Antonio
  Liotta.
\newblock An experimental-based review of image enhancement and image
  restoration methods for underwater imaging.
\newblock \emph{IEEE Access}, 2019.

\bibitem[eriklindernoren(2019)]{code}
eriklindernoren.
\newblock Pytorch-gan.
\newblock \url{https://github.com/eriklindernoren/PyTorch-GAN}, 2019.

\bibitem[Berman et~al.(2020)Berman, Levy, Avidan, and Treibitz]{SQUID}
Dana Berman, Deborah Levy, Shai Avidan, and Tali Treibitz.
\newblock Underwater single image color restoration using haze-lines and a new
  quantitative dataset.
\newblock \emph{IEEE Transactions on Pattern Analysis and Machine
  Intelligence}, 2020.

\bibitem[Shen et~al.(2020)Shen, Gu, Tang, and Zhou]{interfacegan}
Yujun Shen, Jinjin Gu, Xiaoou Tang, and Bolei Zhou.
\newblock Interpreting the latent space of gans for semantic face editing.
\newblock In \emph{Proceedings of the IEEE/CVF Conference on Computer Vision
  and Pattern Recognition}, pages 9243--9252, 2020.

\bibitem[Jiang et~al.(2021)Jiang, Gong, Liu, Cheng, Fang, Shen, Yang, Zhou, and
  Wang]{enlightengan}
Yifan Jiang, Xinyu Gong, Ding Liu, Yu~Cheng, Chen Fang, Xiaohui Shen, Jianchao
  Yang, Pan Zhou, and Zhangyang Wang.
\newblock Enlightengan: Deep light enhancement without paired supervision.
\newblock \emph{IEEE Transactions on Image Processing}, 30:\penalty0
  2340--2349, 2021.

\bibitem[Zhang and Patel(2018)]{did_mdn}
He~Zhang and Vishal~M Patel.
\newblock Density-aware single image de-raining using a multi-stream dense
  network.
\newblock In \emph{Proceedings of the IEEE conference on computer vision and
  pattern recognition}, pages 695--704, 2018.

\bibitem[He et~al.(2011)He, Sun, and Tang]{DCP}
Kaiming He, Jian Sun, and Xiaoou Tang.
\newblock Single image haze removal using dark channel prior.
\newblock \emph{IEEE Transactions on Pattern Analysis and Machine
  Intelligence}, 33\penalty0 (12):\penalty0 2341--2353, 2011.

\end{thebibliography}
}

\end{document}